\documentclass[11pt,a4paper]{article}
\usepackage{times,latexsym}
\usepackage{url}
\usepackage[T1]{fontenc}
\usepackage{amssymb}
\usepackage{amsmath,amsfonts}
\usepackage{svg}
\usepackage{pifont}

\usepackage{booktabs}
\usepackage{multirow}
\usepackage{enumitem}
\usepackage{makecell}
\usepackage{placeins}
\usepackage{subfigure}

\usepackage[acceptedWithA]{tacl2021v1}
\usepackage{tikz}
\usetikzlibrary{positioning}
\usepackage{xspace,mfirstuc,tabulary}
\usetikzlibrary{tikzmark,calc}

\newcommand{\se}[1]{\textcolor{black}{#1}}
\newcommand{\cl}[1]{\textcolor{black}{#1}}
\newcommand{\cll}[1]{\textcolor{black}{#1}}

\newcommand{\ed}[1]{\textcolor{black}{#1}}
\newcommand{\fnl}[1]{\textcolor{black}{#1}}
\newcommand{\reb}[1]{\textcolor{black}{#1}}

\newcommand{\vrobtitle}{CROC}
\newcommand{\metricname}{CROCScore}

\makeatletter
\newcommand{\printCaptionSize}{\texttt{\f@size pt}}
\makeatother

\usepackage{graphicx}
\usepackage{tabularx}
\usepackage{mwe}

\newcommand{\cmark}{\textcolor{green}{\ding{51}}}
\newcommand{\xmark}{\textcolor{red}{\ding{55}}}

\newif\iftaclinstructions
\taclinstructionsfalse 
\iftaclinstructions
\renewcommand{\confidential}{}
\renewcommand{\anonsubtext}{(No author info supplied here, for consistency with
TACL-submission anonymization requirements)}
\newcommand{\instr}
\fi

\iftaclpubformat 

\else

\fi

\definecolor{darkgreen}{rgb}{0.0, 0.5, 0.0}  
\newcommand{\green}[1]{\textcolor{darkgreen}{#1}}
\newcommand{\red}[1]{\textcolor{red}{#1}}

\title{%
  \makebox[\textwidth][c]{%
    \raisebox{-0.54\height}{%
      \includegraphics[width=2.3em]{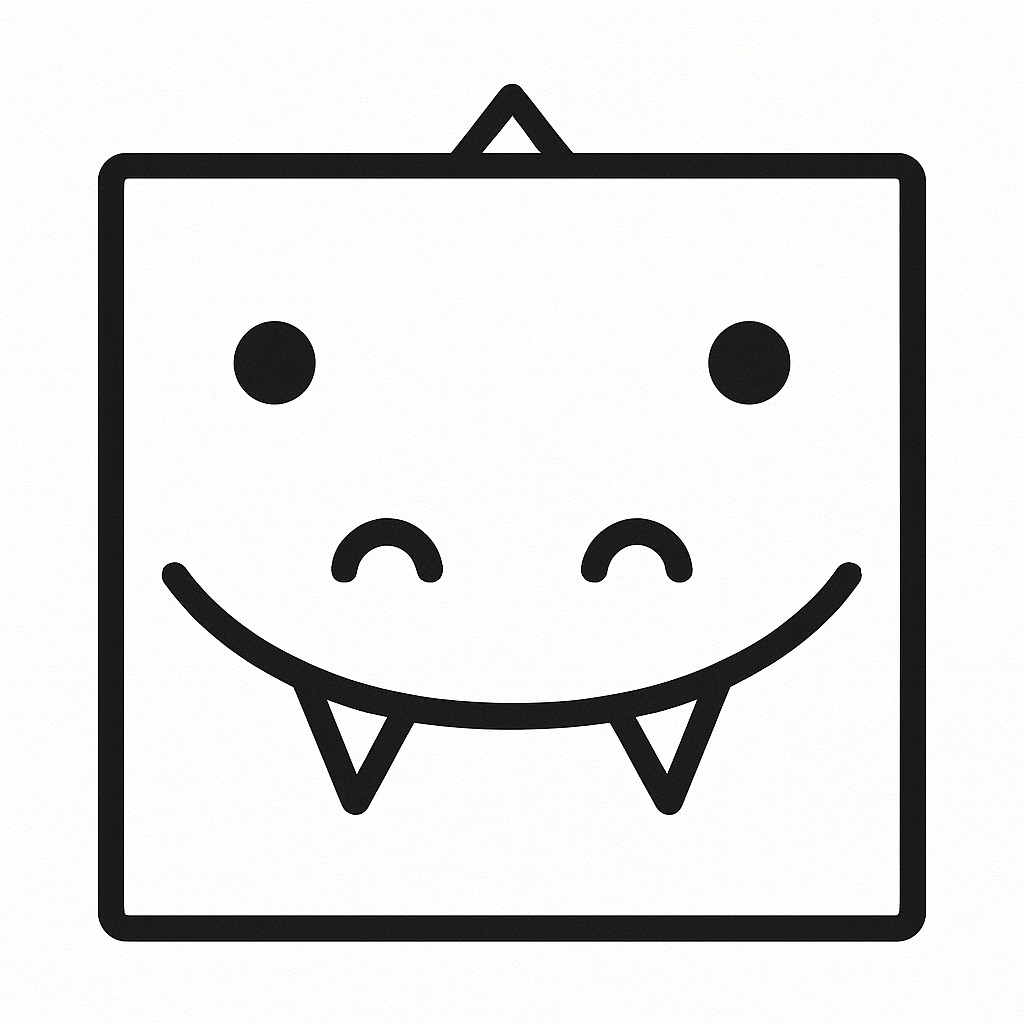}}
    \hspace{0.6em}
    \parbox[t]{0.92\textwidth}{%
      \centering\bfseries
      CROC: Evaluating and Training T2I Metrics\\
      with \ed{Pseudo- and Human-Labeled} Contrastive Robustness Checks}%
  }%
}

\author{
    Christoph Leiter$^{1,2,*,+}$ 
    \and
    Yuki M. Asano$^2$
    \and
    Margret Keuper$^{1,3}$
    \and
    Steffen Eger$^{1,2,+}$\\
    $^1$University of Mannheim,
    $^2$University of Technology Nuremberg, 
    \\
    $^3$Max Planck Institute for Informatics, Saarland Informatics Campus
    \\
    $^*$\texttt{christoph.leiter@uni-mannheim.de}\\$^+$\texttt{NLLG (\url{https://nl2g.github.io/})}
}

\date{}

\begin{document}
\maketitle
\begin{abstract}
The assessment of evaluation metrics (meta-evaluation) is crucial for determining the suitability of existing metrics in text-to-image (T2I) generation tasks. Human-based meta-evaluation is costly and time-intensive, and automated alternatives are scarce. We address this gap and propose \vrobtitle: a scalable framework for \textit{automated \textbf{C}ontrastive \textbf{Ro}bustness \textbf{C}hecks} that systematically probes and quantifies metric robustness by synthesizing contrastive test cases across a comprehensive taxonomy of image properties.
With \vrobtitle, we generate a pseudo-labeled dataset (CROC$^{syn}$) of over 1 million contrastive prompt–image pairs to enable a fine-grained comparison of evaluation metrics.
We also use this dataset to train \metricname, a new metric that achieves state-of-the-art performance among open-source methods, demonstrating an additional key application of our framework. To complement this dataset, we introduce a human-supervised benchmark (CROC$^{hum}$) targeting especially challenging categories. Our results highlight robustness issues in existing metrics: for example, many fail on prompts involving negation, and all tested open-source metrics fail on at least 24\% of cases involving correct identification of body parts.\footnote{Our framework is available at \url{https://github.com/Gringham/CROC/tree/main}.}
\end{abstract}

\section{Introduction}
\label{sec:intro}
\reb{The multimodal task of text-to-image (T2I) generation has advanced rapidly: from text-conditioned GANs \cite[e.g.,][]{pmlr-v48-reed16} to state-of-the-art architectures like diffusion transformers \cite[e.g.,][]{esser2024scalingrectifiedflowtransformers}. T2I models produce images conditioned on textual prompts including desired styles, actions, relationships, or attributes.}

\reb{Evaluating T2I outputs is challenging because multiple images can validly satisfy a prompt, making judgments subjective. Since human evaluation is slow and expensive, automatic metrics are used to assess various quality dimensions \cite{hartwig2025surveyqualitymetricstexttoimage}. 
These metrics rank outputs, benchmark systems, filter training data, and guide fine-tuning and re-ranking \cite{hartwig2025surveyqualitymetricstexttoimage,JMLR:v25:22-0416}. Although numerous evaluation metrics have been proposed, their meta-evaluation (evaluating the metrics themselves) is less researched and typically relies on correlations with human labels \cite[e.g.,][]{Hu_2023_ICCV,cho2024davidsonian,DBLP:journals/corr/abs-2404-16820}. However, this strategy has several shortcomings: (1) it is costly and time-consuming; (2) it offers limited coverage of image properties; (3) older datasets pose a risk of data leakage; and (4) certain correlation measures may unfairly favor metrics \cite{deutsch-etal-2023-ties}.}

\begin{figure}[!ht]
    \centering
    \includegraphics[width=\linewidth]{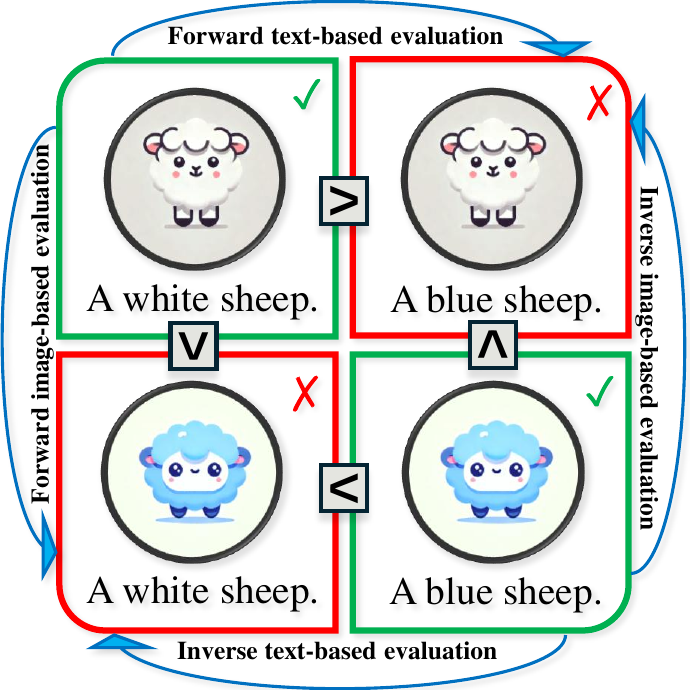}
    \caption{Contrastive evaluation of T2I metrics. Given a text-to-image metric that assigns quality scores to text-image pairs, matching pairs (green) should receive higher scores than non-matching pairs (red). In \textit{text-based} evaluations, the original text is replaced with a contrastive text, while in \textit{image-based} evaluations, the original image is replaced. In \textit{inverse} evaluations, the matching pair is defined by the contrastive text and image used in the \textit{forward} evaluations.}
    \label{fig:comparison_overview}
\end{figure}

\cl{Previous work in machine translation has demonstrated the utility of generated test cases \cite[e.g.][]{sai2021perturbationchecklistsevaluatingnlg, karpinska-etal-2022-demetr, chen-eger-2023-menli} for evaluating the robustness of evaluation metrics. Motivated by their findings, we investigate the viability of automatically generating meta-evaluation datasets for T2I metrics. In particular, we focus on assessing their \textbf{robustness} by identifying which T2I properties and content types each metric handles well and which it does not.}
 To this end, we propose \textbf{\vrobtitle}: automated \textbf{C}ontrastive \textbf{Ro}bustness \textbf{C}hecks (see Figure~\ref{fig:comparison_overview}).
 
\reb{Each \vrobtitle\ sample tests a specific property for text-image alignment (e.g., “can metrics detect the color \textit{white}?”). It contains an original text (e.g., “a white sheep”), a contrastive text (e.g., “a blue sheep”), and one or more images per text. Texts are generated by LLMs and images by diffusion models. Under accurate generation, matching text-image pairs should, by definition, score higher than mismatched ones, thereby minimizing human supervision. Cases that models cannot reliably generate still require human oversight.}

Based on a comprehensive taxonomy of properties, domains, and entities, we construct a large pseudo-labeled (synthetic ground-truth through contrastive generation) dataset, CROC$^{\text{syn}}$, and benchmark recent metrics on it. We also create a human-supervised set, CROC$^{\text{hum}}$, focused on particularly challenging generation categories. As a further use case, we train a new metric, \textbf{CROCScore}, on CROC$^{\text{syn}}$.  
\reb{In summary, we make the following contributions}:
\begin{itemize}
  \item[\green{\checkmark}]
  \textbf{Meta-evaluation framework:}
  \reb{We introduce CROC, automated \textbf{C}ontrastive \textbf{Ro}bustness \textbf{C}hecks, an adaptable framework for T2I metric meta-evaluation. To our knowledge, CROC is the first metric meta-evaluation framework that \emph{jointly} uses LLM-generated, property-wise perturbations and pairwise contrastive checks to verify that metric scores move in the expected direction, reducing human labeling and enabling fine-grained robustness analysis.}

  \item[\green{\checkmark}]
  \textbf{Datasets (CROC$^{\text{syn}}$ and CROC$^{\text{hum}}$):}
  \reb{We construct the large-scale, pseudo-labeled dataset CROC$^{\text{syn}}$, providing $>1$M prompt–image pairs for training and metric comparison. To validate its utility as a benchmark, we conduct a human evaluation and show that evaluated metrics still fall short of human accuracy.
  Further, we curate CROC$^{\text{hum}}$, a human-supervised core set targeting cases that current T2I models struggle to generate.}

  \item[\green{\checkmark}]
  \textbf{CROCScore:}
  \reb{We use CROC$^{\text{syn}}$ to train a new metric CROCScore, achieving state-of-the-art performance among open-source metrics on CROC$^{\text{hum}}$, GenAi-Bench \cite{li2024genaibenchevaluatingimprovingcompositional}, Winoground \cite{thrush_and_ross2022winoground} and TIFA \cite{Hu_2023_ICCV}. This demonstrates another benefit of the pseudo-labeled data.}

  \item[\green{\checkmark}]
  \textbf{Benchmark and robustness findings:}
  \reb{We evaluate 6 metrics on CROC$^{\text{syn}}$ and find that, while VQAScore \cite{lin2024evaluatingtexttovisualgenerationimagetotext} performs best in most evaluation settings, embedding-based metrics like AlignScore \cite{saxon2024who} and BLIP2-ITM \citep{pmlr-v202-li23q} outperform it in some cases, consistent with a previous meta-evaluation \cite{saxon2024who}. On the more challenging CROC$^{\text{hum}}$, performance gaps widen, particularly for image-based evaluation. On this dataset, we additionally evaluate VQAScore with GPT-4o backend (best overall) and our metric CROCScore (best among open-source metrics).
  Fine-grained analyses reveal robustness issues in specific properties. For example, many metrics mishandle negation and all open-source metrics confuse body parts in ca.\ $25\%$ of cases.}
\end{itemize}

\section{Related Work} \label{sec:related}
This section briefly reviews prior work on T2I metrics and their fine-grained meta-evaluation.

\paragraph{T2I evaluation metrics}
The survey by \citet{hartwig2025surveyqualitymetricstexttoimage} groups T2I metrics into two families:
(1) \emph{embedding-based} approaches, e.g., CLIPScore \cite{hessel-etal-2021-clipscore}, measure text-image similarity in a shared embedding space.
(2) \emph{Content-based} approaches, e.g., BVQA \cite{huang2023t2icompbench}, decompose the evaluation process, for example, by reformulating the input text into questions (VQA-based) that a multimodal LLM answers.
In this work, we distinguish \emph{tuned} metrics as a third family:
these metrics, e.g., PickScore \cite{kirstain2023pickapicopendatasetuser}, are trained on human quality judgments. Although most tuned methods build on embedding-based approaches, their training signal and typical failure modes differ.

Several works use these metrics to benchmark \emph{T2I models} with fine-grained prompt suites (e.g., DrawBench and DALL-EVAL \cite{NEURIPS2023_dd83eada,Cho2023DallEval}). They share our emphasis on property-wise prompts, but their goal is model evaluation. Our goal is to \emph{meta-evaluate the metrics} themselves using contrastive checks.

\paragraph{Metric meta-evaluation}
Datasets such as TIFA \cite{Hu_2023_ICCV} provide sample-level, human-labeled scores for text-image pairs.
Recent benchmarks, including Gecko \cite{DBLP:journals/corr/abs-2404-16820} and GenAI-Bench \cite{li2024genaibenchevaluatingimprovingcompositional}, extend this by providing human-labeled scores over diverse sets of images, enabling fine-grained analysis of metric behavior. Our approach instead uses \emph{generated, property-wise contrastive examples}. For each text, we construct matched and contrasted pairs that differ on a single property, enabling targeted robustness analyses while minimizing manual labeling because labels are implied by construction.

T2IScoreScore \cite{saxon2024who} evaluates whether metric scores decrease along controlled degradations that are injected into aligned pairs. Our property-based perturbations target fine-grained behavior across specific properties, which is enabled through the contrastive design.

Winoground \cite{thrush_and_ross2022winoground} and follow-up benchmarks like Winoground-T2I \cite{zhu2023contrastivecompositionalbenchmarktexttoimage} also use contrastive setups to probe compositionality of multimodal models. The latter includes a small-scale meta-evaluation via correlation with human Likert scores. In contrast to these, we set a clear focus on the fine-grained, property-wise meta-evaluation of T2I metric robustness through generated, pseudo-labeled data.

Compared to the above benchmarks, CROC$^{\text{hum}}$ also contains more human-verified text-image pairs and CROC$^{\text{syn}}$ is on a much larger scale (see Table \ref{tab:sizeComparison}). We also introduce CROCScore, a strong open-source metric trained on CROC$^{\text{syn}}$.

\begin{table}[t]
\centering
\setlength{\tabcolsep}{6pt}
\begin{tabular}{l|r|r}
\toprule
Benchmark & Texts & T-I Pairs\\
\midrule
TIFA & 160 & 800  \\
Gecko & 2k  & 8k \\
GenAI-Bench  & 1{,}6k & 12{,}8k \\
T2IScoreScore & 165 err.\ graphs & 2{,}8k \\
Winoground & 800 & 1{,}6k \\
Win.\-T2I (S-100) & 200 & 800 \\
CROC$^{\text{hum}}$ & 560 & \(\sim\)23{,}8k \\
CROC$^{\text{syn}}$ & \(\sim\)220k & \(>\)1M \\
\bottomrule
\end{tabular}
\caption{Comparison of related dataset sizes.}
\label{tab:sizeComparison}
\end{table}

More broadly, our work relates to metrics that exploit contrasts in model representations \cite[e.g.,][]{wang2025contrastscorehigherqualitybiased} and to contrastive pre-training of joint text-image encoders \cite[e.g.,][]{pmlr-v139-jia21b}. We differ by using a fine-grained generated dataset to meta-evaluate and fine-tune T2I metrics.

\section{Methodology}
\label{sec:methodology}
In this section, we describe the evaluation setup and generation framework of \vrobtitle\ and the training approach of CROCScore. 

\begin{figure*}[htbp]
  \includegraphics[width=\linewidth]{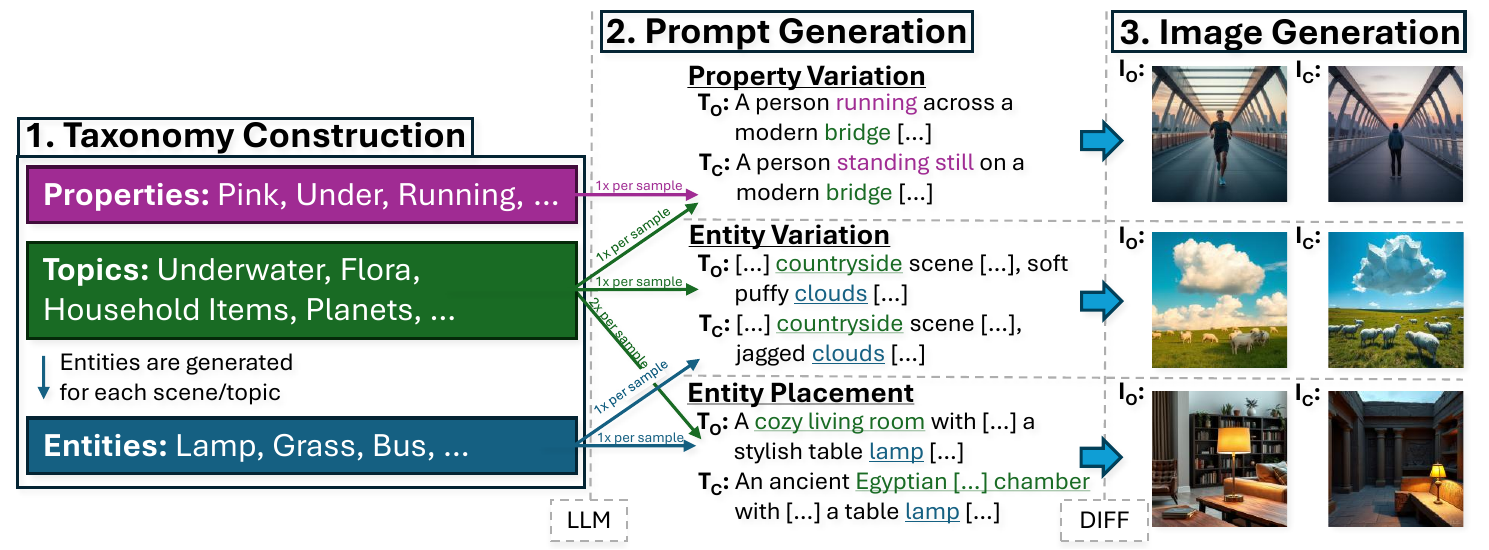}
  \caption{\reb{Overview of the CROC$^{\text{syn}}$ data generation process. First, we construct our taxonomy based on related work and with LLM support. The image shows example categories. Second, we pass selected categories of the taxonomy to an LLM (e.g., ``running'' and ``bridges and infrastructure'') to generate the original text $T_O$ and the contrast text $T_C$ for one sample. Then these are passed to a diffusion model to generate the original image $I_O$ and the contrast image $I_C$. 
  }}
  \label{fig:data_generation}
\end{figure*}

\subsection{Evaluation directions}
\label{evalsetups}
\cl{A T2I model $G(T)=I$ generates an image $I$ based on an image generation prompt $T$. Accordingly, a} T2I metric \cl{$M(T, I)=s$} assigns a score $s\in\mathbb{R}$ that indicates how well $I$ follows $T$ (i.e., their alignment, higher is better). Further, we refer to a $T$-$I$ pair where $I$ correctly follows $T$ as a \textbf{matching pair}. Likewise, a pair where $I$ does not follow $T$ is referred to as a \textbf{contrast pair}.

Inspired by related work in the domain of MT evaluation \cite{chen-eger-2023-menli}, \cl{we meta-evaluate the robustness of T2I metrics in a contrastive setup (see Figure \ref{fig:comparison_overview}). That means, we test whether the condition 
\begin{align}
    \ed{M(\text{matching pair}) > M(\text{contrast pair})}
\end{align} is correctly fulfilled.} 
Specifically, we design examples with two contrasting prompts $T_O$ and $T_C$, as well as images $I_O$ and $I_C$ \cll{(\textbf{O} for \textbf{O}riginal and \textbf{C} for \textbf{C}ontrast)}. Here, $T_O$-$I_O$ and
$T_C$-$I_C$ are matching pairs, while $T_O$-$I_C$ and $T_C$-$I_O$ are contrast pairs. For example, in Figure \ref{fig:comparison_overview} the green cells show matching $T$-$I$ pairs (e.g., the text ``A white sheep'' and the image of a white sheep) that should receive higher metric scores than the red cells that show contrast pairs (e.g., the text ``A white sheep'' and the image of a blue sheep). \cll{Notably, by controlling $T_O$ and $T_C$ so that they differ solely in specific features (such as color or position), we facilitate fine-grained robustness tests.}
\cl{There are four possible evaluation directions:}

\cl{\textbf{\reb{1.} Image-based} evaluations compare a matching pair to a contrast pair that changes the image (column-wise comparison in Figure \ref{fig:comparison_overview}). For example, we can evaluate whether $\cl{M}(T_{O},I_{O})>\cl{M}(T_{O},I_{C})$ is true. This evaluation corresponds to the question: ``Which image was more likely generated from $T_{O}$?'' }

\cl{\textbf{\reb{2.} Text-based} evaluations compare a matching pair to a contrast pair that changes the text (row-wise comparison in Figure \ref{fig:comparison_overview}).
For example, \cl{we can evaluate whether} 
\cl{$M(T_O,I_O)>M(T_C,I_O)$} is true. This evaluation corresponds to the question: ``Which prompt is more likely to have generated $I_O$?'' \cll{This evaluation type is related to image captioning. The difference is that the images were generated from the prompt and not vice versa.}} 

\cl{\textbf{\reb{3.\ + 4.}} In our prompt creation process (see \cll{\S}\ref{subsec:contrastiveDatasetGeneration}), we create $T_C$ by changing an \textit{original} prompt $T_O$. Therefore, $T_{C}$ is more likely to feature unusual scenarios. When the matching pair of the comparison is $T_{O}$-$I_{O}$, we refer to the evaluation as \textbf{forward evaluation}. In contrast, if the matching pair is $T_{C}$-$I_{C}$, we refer to the evaluation as \textbf{inverse evaluation}.}

\subsection{Contrastive Dataset Generation}
\label{subsec:contrastiveDatasetGeneration}
\reb{We propose two complementary dataset-construction pipelines: (1) a \textit{pseudo-labeled} pipeline that generates \textbf{CROC$^{\text{syn}}$}, a large-scale dataset designed to broadly cover text-to-image (T2I) alignment use cases; and (2) a \textit{human-supervised} pipeline that curates \textbf{CROC$^{\text{hum}}$}, a suite of evaluation tests targeting especially challenging properties of image generation.}

\paragraph{\cll{Pseudo-labeled} data generation} \reb{The \textit{pseudo-labeled} pipeline (see Figure \ref{fig:data_generation}) creates comprehensive contrastive samples ($T_O$, $T_C$, $I_O$, $I_C$) in three steps: (1) \textit{taxonomy construction}, (2) \textit{text generation} and (3) \textit{image generation}. The taxonomy is hierarchical and contains 64 image properties (e.g., relations, attributes, colors), 51 topics (e.g., nature, people, architecture) and 158 entities (e.g., eagle, bridge, bus). Each element is paired with a short textual description. The taxonomy of properties, example topics and example entities is shown in Appendix \ref{app:taxProperties}.} To construct this taxonomy, we first consolidated initial properties from \citet{DBLP:journals/corr/abs-2404-16820}, \citet{hartwig2025surveyqualitymetricstexttoimage}, \citet{foote_design_guides} and \citet{chen-eger-2023-menli}. \reb{Then, we manually define and interactively generate (with GPT-4o) a broad hierarchy of further subclasses, topics and respective descriptions. Entities are created for every topic leaf node, e.g., \textit{grass} for \textit{flora} and \textit{deer} for \textit{fauna}.}
Based on this taxonomy, we create three \reb{types of test prompts} \ed{(one that is based on properties and two that are based on entities)}: 

\textbf{1. Property variation:} \cl{Here, $T_O$ features \reb{a} property selected from the taxonomy and $T_C$ strongly changes this property. Referring back to Figure \ref{fig:comparison_overview}, the property \textit{white} in $T_O$ ``A white sheep'' is changed to \textit{blue}. 
Property variation is the main test \reb{type in} our robustness evaluation, as it allows for a fine-grained check of metric capabilities.} 

\textbf{2. Entity placement:} \cl{In an initial experiment, we generated contrastive text-image pairs from T2ICompBench \cite{huang2023t2icompbench} and found that human annotators can rate inverse evaluation directions (with unexpected prompts) better than metrics (see Table \ref{tab:t2icompbench}). Motivated by this, we evaluate the capability of metrics to correctly rate unexpected matching text-image pairs higher than expected contrast pairs\footnote{This is to some degree also inherent to \textit{property variation}, because our prompt generation process \reb{prompts an LLM} to first construct the original text and then change something. But due to the large variety of tested properties, both texts may be unexpected\se{.}}. Here, $T_O$ describes an entity with its \reb{source topic, i.e., where it naturally occurs}. For example, this could be a \textit{sheep} on a \textit{field}. $T_C$ then places the entity in an unusual topic, e.g., a \textit{sheep} in a \textit{city}.}

\textbf{3. Entity variation:} \cl{As another way of testing the possible unexpectedness bias of T2I metrics, we generate $T_O$ such that an entity is described naturally. Then, $T_C$ describes the entity with an altered description. For example, this could be ``a sheep with two eyes'' vs.\ ``a sheep with three eyes''.} 

\reb{In the final step, the images $I_O$ and $I_C$ are generated for $T_O$ and $T_C$, respectively.}

\paragraph{Practical considerations:  CROC$^{\text{syn}}$} 
\cl{In practice, we set $I_O=G(T_O)$ and $I_C=G(T_C)$. However, just like T2I evaluation metrics, the \fnl{text generation models and} T2I models are imperfect and may not correctly follow the prompt. This presents a chicken-and-egg problem: normally, T2I evaluation metrics evaluate T2I models. In our setup, we meta-evaluate T2I evaluation metrics \textbf{with} T2I models. T2I evaluation metrics 
are not perfect \cll{and their quality can be quantified} in terms of human correlation. Similarly, our automatic evaluation setup for T2I evaluation metrics is not perfect. Therefore, (1) we measure its quality with human annotations (see \S \ref{sub_sec:human_eval}), \cll{and} 
(2) \se{w}e compare the results with CROC$^{\text{hum}}$ 
 (see \S \ref{subsec:contrastiveDatasetGeneration}).} 
\se{Further,} 
(3) \cll{f}or each prompt \cll{($T_O$ resp.\ $T_C$)}, we generate  
\se{$n>1$}
images \cll{($I^{1,\ldots,n}_O$ resp.\ $I^{1,\ldots,n}_C$).} 
\cll{Then, f}or the \cll{\textbf{forward text-based}} setups, we evaluate: 
\begin{align}
j^* = \underset{i=1,\ldots,n}{\operatorname{argmax}}\, \cl{M}(\cll{T_O, I^i_O}),\nonumber\\
\cl{M}(\cll{T_O, I^{j^*}}) > \cl{M}(\cll{T_C, I^{j^*}})
\end{align}
where $i$ and $j^*$ are indices for the images. 
\cll{That means, if at least one image $I^{j^*}_O$ follows $T_O$ correctly, the metric should correctly pick that image (assign the highest score to $M(T_O, I^{j^*})$), otherwise it is an error of the metric and not an error of the setup (
\se{see} 
Appendix \ref{DetailedExamples} for a detailed example). \reb{For $n=5$ images\se{,} this setup raises the accuracy of a random-scoring baseline to $\frac{5}{6}\approx 83.\overline{3}\%$, since among $n{+}1$ i.i.d.\ draws (i.e., 5 options from the matching side, given through the maximum, and 1 option from the non-matching side) each is equally likely to be the maximum, giving $1-\frac{1}{n+1}$.}
We use this later to scale the accuracies for comparability.} \cll{For the \textbf{forward image-based setup}\se{,} we compare}:
\begin{equation}
\underset{i=1,\ldots,n}{\operatorname{max}}\, \cl{M}(\cll{T_O, I^i_O}) > \underset{j=1,\ldots,n}{\operatorname{max}}\, \cl{M}(\cll{T_O, I^j_C})
\end{equation}
\cll{That means, the $I^i_O$ that matches $T_O$ best should be rated higher than the $I^j_C$ that matches $T_O$ best. \ed{Here, the method is different because for text-based evaluation we have many images per prompt. On the other hand, for image generation, we often do not have multiple prompts per image because the prompt generation is less restricted \reb{and prompts can contain different content even though they have the same topic and property}.} For \textbf{inverse} evaluations\se{,} $T_O$ is swapped with $T_C$ and $I^i_O$ is swapped with $I^i_C$ (see Appendix \ref{inverseEquations}). For this setup\se{,} the accuracy of a random baseline is 50\%. While there is no guarantee that every single sample is generated correctly, these considerations ensure that evaluation on our dataset provides meaningful fine-grained comparisons between metrics.}
\reb{Another circularity that may occur is that, just like LLM-based evaluators can have biases that favor their own generations \cite{NEURIPS2024_7f1f0218}, the mLLM-based metrics that we are testing may be biased if the data-generation LLMs are related to the metric backbone. We argue that this effect should be partly mitigated, because bias in a contrastive setup would affect the rating of both, matching and non-matching text-image pairs.\footnote{As a further mitigation, in our experiments we only select generation LLMs from different model families than the metric-backbones. Also, we do not evaluate CROCScore on the dataset it was trained on (CROC$^\text{syn}$).}}

\paragraph{Human-supervised data generation}\label{sec:hum_sup_gen}
\reb{Supplementary to the pseudo-labeled pipeline of CROC$^{\text{syn}}$, we construct the human-supervised dataset, CROC$^{\text{hum}}$. It addresses selected categories that are especially challenging for T2I models. These categories are critical because evaluation metrics should remain reliable when T2I models fail. Since learned metrics often differ from T2I models in architecture and training objectives, it is not obvious that they fail on the same instances. CROC$^{\text{hum}}$ allows us to probe where metrics and the models they evaluate share weaknesses.
It is built in a five-step process:
(1) \textit{category selection}, (2) \textit{template-based prompt construction}, (3) \textit{grammar check}, (4) \textit{image generation}, and (5) \textit{human validation}.}

\reb{In \textit{step one}, we \textit{choose eight challenging categories} (see Appendix \ref{app:human_categories}).} The categories \textit{body parts} and \textit{parts of things} are motivated by persistent difficulties many T2I models have with rendering hands \cite[e.g.,][]{Narasimhaswamy_2024_CVPR}. 
Further, the categories \textit{counting}, \textit{shapes}, \textit{size relation}, \textit{spatial relation}, \textit{action} and \textit{negation}  
were \se{already} included in prior correlation-based benchmarks \cite[e.g.,][]{DBLP:journals/corr/abs-2404-16820, li2024genaibenchevaluatingimprovingcompositional}. 
We differ in our contrastive setup and in our prompt construction. 

\reb{In \textit{step two}, we generate original prompts $T_O$ and contrast prompts $T_C$ for each selected category.} For action, body parts and parts of things, we interactively create $T_O$ and $T_C$ with GPT-4o\cite{gpt4o}. For example, we use colors to change a highlighted body part between $T_O$ (e.g., \textit{A red ring finger}) and $T_C$ (e.g., \textit{A red thumb}).
For the remaining five categories, we randomly select one or two entities from the CROC$^\text{syn}$ taxonomy and one property from a predefined list, for example, the entities \textit{person} and \textit{car}, and the property \textit{left of} for spatial relations. We then fill fixed templates to form simple prompts such as ``a person left of a car'' or ``a person and no car'' and use GPT-4o to verify the grammar (\textit{step three}). A detailed example \cll{for \textit{body parts}} is described in Appendix \ref{DetailedExamples} (example 2). Once all prompts are generated, we generate $I_O$ and $I_C$ (\textit{step 4}). \ed{Negation is a special category, because we can use a trick to generate high quality $I_C$ images: instead of generating an image of $T_C=$``A and not B'', we simply generate an image of the alternative prompt ``A'' that does not contain the negation, but still compare $T_C$ during the evaluation.}

\reb{In the \textit{final step}, we perform human verification of every text-image pair. To facilitate simple and fast verification of a large number of examples, we (1) construct short prompts\footnote{In CROC$^{\text{syn}}$, we opt for long, descriptive prompts, to increase the quality and diversity of the generated images. In contrast, in CROC$^{\text{hum}}$, we opt for short prompts, which might reduce image quality, but allow human annotators to verify the T2I alignment of the images with more certainty in a shorter amount of time.}, (2) generate a large number of images (in our experiments $n=100$) per prompt and (3) verify by filtering, i.e., annotators see the prompt and all images in a scrollable list (e.g., a file explorer) and remove all images that do not fit the respective text. The exact annotation setup is described in \S\ref{dataset_setups}.}

\paragraph{Practical considerations:  human-sup.\ dataset}
Due to the human verification through filtering, CROC$^{\text{hum}}$ has strong prompt–image alignment, so it is unnecessary to apply the same selection strategies used for CROC$^{\text{syn}}$. However, CROC$^{\text{hum}}$ contains a varying number of images per prompt \ed{(varying $i$ and $j$)}. Therefore, we treat all prompts equally by averaging the sample-wise performance.
For the forward text-based evaluation, for each prompt $T_O$, we compute the ratio of cases where $M(T_O, I^i_O) > M(T_C, I^i_O)$ across all $i$.
For the forward image-based evaluation, we compute the ratio of cases where $M(T_O, I^i_O) > M(T_O, I^j_C)$ across all $i,j$.
The final performance for a category is then obtained by averaging these ratios across prompts.
For inverse evaluation, we swap $O$ and $C$ accordingly.

\paragraph{CROCScore}
The basis for CROCScore is inspired by VQAScore \cite{lin2024evaluatingtexttovisualgenerationimagetotext}, which prompts a multimodal LLM (mLLM) to answer a simple question like ``Does this image show \{prompt\}'' and uses the probability of the answer ``Yes'' as a metric score. We extend this approach and ask a question like ``Does this image show the following content:'\{prompt\}'? Answer with Yes or No.''. Then the score is computed as $p(\text{Yes})-p(\text{No})$. \ed{This matches our contrastive setup, in which non-matching pairs should have a high No-probability (and a low Yes-probability).} During training, for each sample in ou\se{r}  dataset, we randomly select either the matching or contrast pair and further fine-tune a\se{n} \se{mLLM} to generate \textit{Yes} for matching pairs and \textit{No} for contrast pairs.

\section{Experiment Setup} \label{sec:setup} \paragraph{Models and infrastructure} We run computations on a Slurm cluster with Nvidia A40 and A100 graphics cards \cll{(Table \ref{metricConfig} compares metric runtimes)}. For \cl{\cll{pseudo-labeled}} prompt generation, we use the two models\se{,} DeepSeek-R1-Distill-Qwen-14B \cite{deepseekai2025deepseekr1incentivizingreasoningcapability} and QwQ-32B \cite{qwq32b}, because of their strong performance on text generation leaderboards and relative cost-effectiveness. The runtime is ca.\ three hours per model (2 resp.\ 4 GPUs). For image generation, we use FLUX.1-schnell \cite{flux2024} and Stable-diffusion-3.5-large-turbo \cite{stability2024} that are fine-tuned to require fewer generation steps (with the trade-off of lower generation quality) to ease the computational requirements. \cll{The image generation took an average of ca.\ 5h on 120 GPUs.}

\paragraph{T2I metrics} We compare several classes of T2I metrics \cl{(see \S\ref{sec:related} for details on the metrics and Table \ref{metricConfig} for their configuration)}. We use the \textbf{embedding-based metrics} CLIPScore \cite{hessel-etal-2021-clipscore} and AlignScore \cite{saxon2024who}. \cll{We also evaluate BLIP2-ITM, which was trained with an image-text matching objective \citep{pmlr-v202-li23q}.} Further, we test the \textbf{trained} metric PickScore \cite{kirstain2023pickapicopendatasetuser} and the more recent \textbf{visual question-answering (VQA) metrics} VQAScore \cite{lin2024evaluatingtexttovisualgenerationimagetotext} and BVQA \cite{huang2023t2icompbench}. \cll{On the human-supervised dataset, we also evaluate VQAScore with a closed-source GPT-4o backend. \ed{We only evaluate it here because of its costly inference (ca.\ \$150 on the supervised dataset).}}

\begin{figure*}[htbp]
  \centering
  \noindent
  \begin{minipage}[t]{0.47\textwidth}
    \centering
    \textbf{Text-based Evaluation}\\[-0.5em]
    \begin{minipage}[t]{0.32\linewidth}
      \centering
      \footnotesize \textbf{FC: AlignScore}\\[-\baselineskip]
      \tikz[baseline=(current bounding box.north)]
        \node {\includegraphics[width=0.96\linewidth]{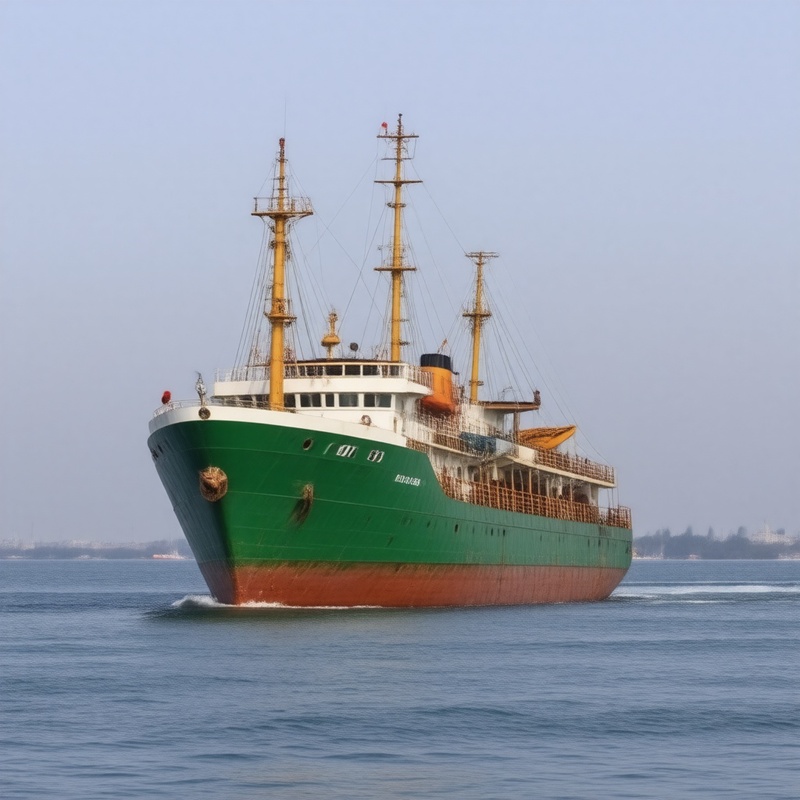}    }
             node[anchor=north east, xshift=33pt, yshift=33pt,
                  text=green, fill=white, draw=black, line width=0.8pt, inner sep=1.5pt] {\cmark};\\
      \footnotesize \textbf{C}: Parts of Things \\
      \footnotesize \textbf{\cmark}: \reb{A ship with a green hull.} \\
      \footnotesize \textbf{\xmark}: A ship with a green mast.  \\
    \end{minipage}\hfill
    \begin{minipage}[t]{0.32\linewidth}
      \centering
      \footnotesize \textbf{FC: PickScore}\\[-\baselineskip]
      \tikz[baseline=(current bounding box.north)]
        \node {\includegraphics[width=0.96\linewidth]{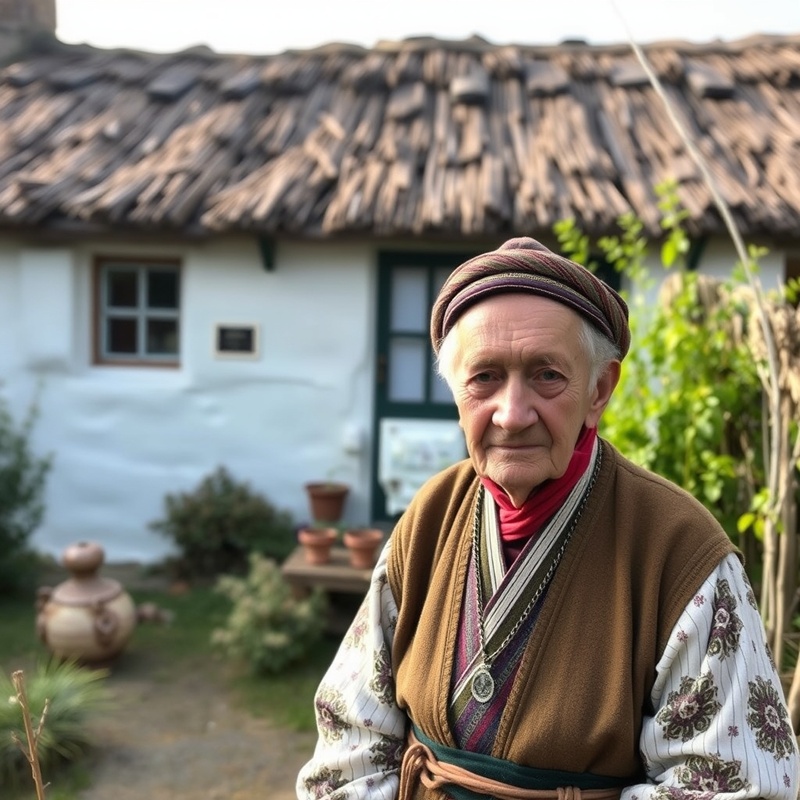}}
             node[anchor=north east, xshift=33pt, yshift=33pt,
                  text=green, fill=white, draw=black, line width=0.8pt, inner sep=1.5pt] {\cmark};\\
      \footnotesize \textbf{C}: Negation \\
      \footnotesize \textbf{\cmark}: \reb{A cottage and an elder.} \\
      \footnotesize \textbf{\xmark}: A cottage and no elder.  \\
    \end{minipage}\hfill
    \begin{minipage}[t]{0.32\linewidth}
      \centering
      \footnotesize \textbf{FC: BVQA}\\[-\baselineskip]
      \tikz[baseline=(current bounding box.north)]
        \node {\includegraphics[width=0.96\linewidth]{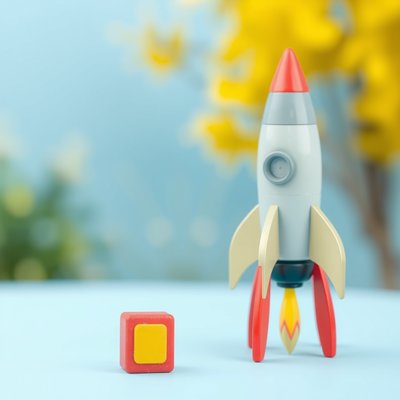}}
             node[anchor=north east, xshift=33pt, yshift=33pt,
                  text=green, fill=white, draw=black, line width=0.8pt, inner sep=1.5pt] {\cmark};\\
      \footnotesize \textbf{C}: Size Relation \\
      \footnotesize \textbf{\cmark}: A smaller square and a bigger rocket.\\
      \footnotesize \textbf{\xmark}: A bigger square and a smaller rocket. \\
    \end{minipage}
  \end{minipage}%
  \quad\vrule width 0.2pt\quad
  \begin{minipage}[t]{0.47\textwidth}
    \centering
    \textbf{Image-based Evaluation}\\[-0.5em]
    \begin{minipage}[t]{0.32\linewidth}
      \centering
      \footnotesize \textbf{FC: BLIP2-ITM}\\[-\baselineskip]
      \tikz[baseline=(current bounding box.north)]
        \node {\includegraphics[width=0.96\linewidth]{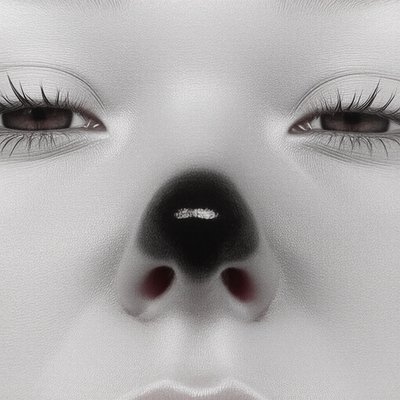}}
             node[anchor=north east, xshift=33pt, yshift=33pt,
                  text=green, fill=white, draw=black, line width=0.8pt, inner sep=1.5pt] {\cmark};\\
      \tikz[baseline=(current bounding box.north)]
        \node {\includegraphics[width=0.96\linewidth]{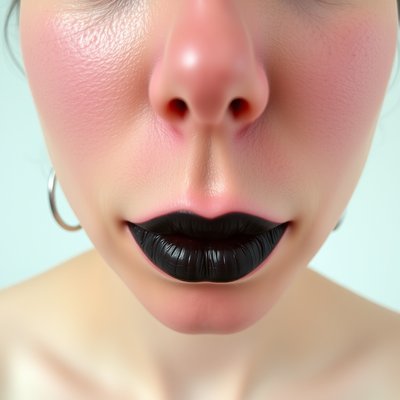}}
             node[anchor=north east, xshift=33pt, yshift=33pt,
                  text=red, fill=white, draw=black, line width=0.8pt, inner sep=1.5pt] {\xmark};\\
      \footnotesize \textbf{C}: Body Parts \\
      \footnotesize \cmark: A face with only its nose colored black. \\
    \end{minipage}\hfill
    \begin{minipage}[t]{0.32\linewidth}
      \centering
      \footnotesize \textbf{FC: VQAScore}\\[-\baselineskip]
      \tikz[baseline=(current bounding box.north)]
        \node {\includegraphics[width=0.96\linewidth]{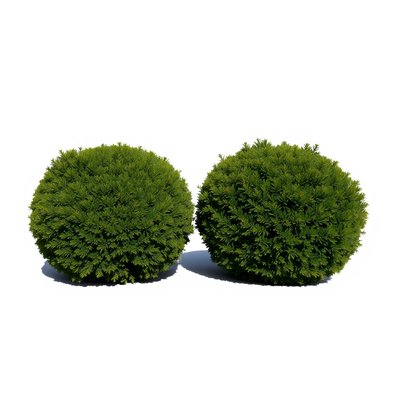}}
             node[anchor=north east, xshift=33pt, yshift=33pt,
                  text=green, fill=white, draw=black, line width=0.8pt, inner sep=1.5pt] {\cmark};\\
      \tikz[baseline=(current bounding box.north)]
        \node {\includegraphics[width=0.96\linewidth]{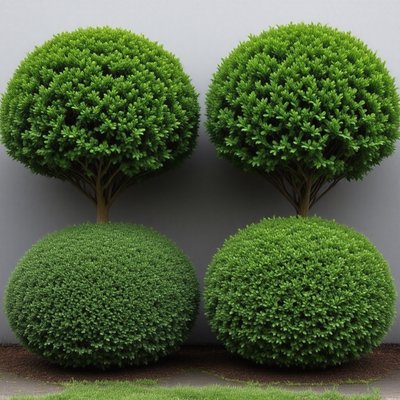}}
             node[anchor=north east, xshift=33pt, yshift=33pt,
                  text=red, fill=white, draw=black, line width=0.8pt, inner sep=1.5pt] {\xmark};\\
      \footnotesize \textbf{C}: Counting \\
      \footnotesize \cmark: Two bushes. \\
    \end{minipage}\hfill
    \begin{minipage}[t]{0.32\linewidth}
      \centering
      \footnotesize \textbf{FC: CLIPScore}\\[-\baselineskip]
      \tikz[baseline=(current bounding box.north)]
        \node {\includegraphics[width=0.96\linewidth]{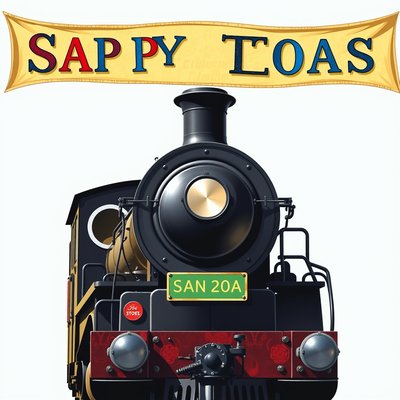}}
             node[anchor=north east, xshift=33pt, yshift=33pt,
                  text=green, fill=white, draw=black, line width=0.8pt, inner sep=1.5pt] {\cmark};\\
      \tikz[baseline=(current bounding box.north)]
        \node {\includegraphics[width=0.96\linewidth]{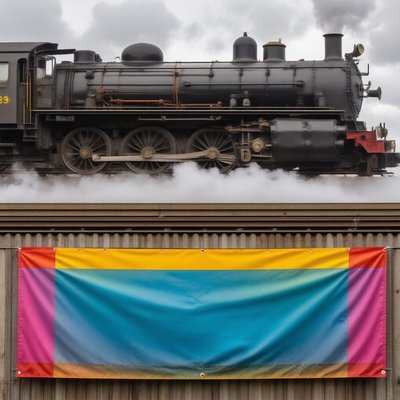}}
             node[anchor=north east, xshift=33pt, yshift=33pt,
                  text=red, fill=white, draw=black, line width=0.8pt, inner sep=1.5pt] {\xmark};\\
      \footnotesize\textbf{C}: Spatial Rel.\ \\
      \footnotesize \cmark: A banner above a steam locomotive.\\
    \end{minipage}
  \end{minipage}

  \caption{Selected metric failure-cases CROC$^{\text{hum}}$. \textit{FC} shows a metric that failed on this example and \textit{C} shows the category of the example.  \cmark{} indicate\se{s} the matching text-image pair. For text-based evaluation, the metric falsely rates the text with \xmark\ \se{higher} than the text with \cmark. For image-based evaluation, the metric falsely rates the image with \xmark\ higher than the image with \cmark.}
  \label{fig:combined_six_columns}
\end{figure*}

\paragraph{The \vrobtitle\ Datasets}\label{dataset_setups} \cl{For \textbf{CROC$^{\text{syn}}$}, we generate up to 20 prompt-contrast pairs per \textit{input description} (five for each LLM-T2I model combination).\footnote{If the output is not in a valid format, we drop it. We choose JSON for simple output parsing.}} 
As we described earlier, we place each prompt type (property variation, entity variation and entity placement) with one of 51 topics (e.g., medieval, landscapes). \cll{Thus, the \textit{input description} can be one of three types: (1) The first type is a property variation pair; for these, we generate all 51 combinations. (2) The second type is an entity variation sample; here, we choose ten random topics for each entity. (3) The third type is an entity placement sample; in this case, we select ten random combinations of topic and alternative topic for each entity.}
For example, for entity placement, we choose ten setups that place the entity ``eagle'' from a natural topic in a randomly chosen alternative topic. In total, CROC$^{\text{syn}}$ contains 25,693 unique entity variation, 28,723 unique entity \reb{placement} and 56,984 unique property variation prompt-contrast pairs. Further, we generate \cl{$n=5$} different images per prompt to increase the robustness of our setup. Appendix \ref{DetailedExamples} shows a full example for one property. \cll{The prompts in this setup have a maximum token count of 180. The prompting guides that we use in the prompt templates suggest detailed prompts to increase the quality of generated images. However, the generated prompts can be long and difficult to \ed{process} in human experiments. \ed{Also, CLIP-based models like CLIPScore have a disadvantage on long prompts with more than 77 tokens, where they are cut off (see Figure \ref{fig:croc-wide} for an analysis).}}

For \textbf{CROC$^{\text{hum}}$}, we create 280 prompts and respective contrast prompts (four test categories with 50 prompts and four with 20 prompts). For each prompt and contrast prompt, we create 50 images with the Stable Diffusion model and 50 images with the Flux model. \reb{This means, in total, we create 56{,}000 text-image pairs. \reb{In Appendix \ref{DetailedExamples} we show a full example of the human-supervised data creation}. 
As described in \S\ref{sec:hum_sup_gen}, we manually filter out images that do not fulfill the category we are evaluating}. For example, for the \reb{category} ``counting'' and a prompt ``four fingers'', we manually remove images that do not show exactly four fingers. 
\reb{ 
During this human verification, we keep images with visual distortions if the matching property is fulfilled and the non-matching one is not fulfilled. 
For example, for ``part of things'' prompts about \textit{calculators}, we accept images of calculators that show incorrect button ordering if the contrasting prompts ``a blue button'' vs.\ ``a blue display'' are fulfilled. For the prompt ``a blue 9-button'', we only accept images that verifiably indicate the button through text or position. For spatial relations, like ``left of'', we consider the position of the viewer.} 

As we deliberately choose hard cases for image generation, some prompts do not generate any valid image. \reb{For these, we augment our data to have at least three valid images by interactively generating matching images with GPT-4o (for body parts) and GPT5 \cite{GPT5} (others) in multi-turn conversations}. 
After filtering, the dataset contains 12{,}090 Flux and 11{,}497 Stable Diffusion images out of 56,000 images \ed{(+245 GPT images)}. \reb{This means, over half of the images are removed, due to unsuccessful generation for these challenging categories.} The filtering was conducted by one person with fluent English skills in approximately 25 hours. While this may introduce a degree of subjectivity, for most prompts it is resolved through the contrastive setup: even if a match is subjective, it will be more matching than the contrast sample.
We further verify the validity of CROC$^{\text{hum}}$ with a \textbf{human evaluation} in \S\ref{sub_sec:human_eval}. Here, 3 annotators annotate the same 480 examples (60 per category, stratified across evaluation direction) and have to select matching text-image pairs over non-matching ones. Notably, because we use 280 prompts and contrast prompts these 480 samples cover at least one original or contrasting prompt of every data sample. 

\paragraph{Human evaluation of CROC$^{\text{syn}}$}
\label{human_eval} 
To assess the quality of the pseudo-labeled generation process, we conduct a human study on CROC$^{\text{syn}}$ using Prolific annotators.\footnote{\url{www.prolific.com}. We target fluent English speakers residing in the UK or USA and, for $\sim$80\% of tasks, require at least a master’s degree.} For every annotation, workers are presented with one text ($T_O$ or $T_C$) and five images ($I_O^{\{1,\ldots,5\}}$ or $I_C^{\{1,\ldots,5\}}$). Their task is to assign a continuous 1–5 alignment rating (slider with two decimal digits) that indicates how well the best matching image is aligned with the text (1 indicates no alignment, while 5 indicates perfect alignment). We choose continuous scores to reduce the amount of ties among scores for low or high quality images. This setup mirrors the max-over-images operation used in our image-based aggregation.
We select 500 samples that are stratified across properties, entities, text- and image-generation models. For every sample, we evaluate all four text-image directions: (1) $T_O$–$I_O^{\{1,\ldots,5\}}$, (2) $T_C$–$I_C^{\{1,\ldots,5\}}$, (3) $T_O$–$I_C^{\{1,\ldots,5\}}$, and (4) $T_C$–$I_O^{\{1,\ldots,5\}}$, yielding 2{,}000 text-image pairs (each consists of one text and five candidate images). We split the 2{,}000 pairs into 26 batches. Each text-image pair receives scores from three Prolific workers (amounting to 77 annotators, because one annotator annotated two batches). Within each batch, the four image-pairs (one per direction) of the same sample are shown in random order. In addition, a subset of 380 pairs is labeled by three in-house annotators and analyzed separately.
For aggregation, we normalize ratings per annotator via z-scoring (subtract the mean and divide by standard deviation) and average the normalized scores to obtain a single score for each item–direction. We then compute the four evaluation directions identically to the automatic metrics. The total annotation cost is ca.\ \$1025 with annotation times ranging between 20min.\ up to 1h20min.\ per batch. To increase annotation quality, ca.\ every 12 samples we display a simple attention check asking to select a score from a certain range hidden within the text.

\paragraph{Training and evaluating CROCScore}
To train CROCScore, we fine-tune the vision encoder and LoRA of phi4-multimodal-instruct \cite{phimodel} on two H100 GPUs. We first select a subset of CROC$^{\text{syn}}$ that includes three data samples for every property variation-topic combination and a total of 200 entity variation prompts. We do not include entity placement, as our analysis in \S\ref{quant_res} shows that metrics already perform strongly on it. Further, we restrict the dataset to contain only DeepSeek and Flux generated text-image pairs to reduce potentially confounding prompting styles. For each data point, we include two random evaluation directions, where the overall distribution is 40\% matching and 60\% non-matching with the intention to make the model slightly more critical towards errors. For the training, we use 12k text-image pairs from this data sample. Detailed parameters can be found in Appendix \ref{app:TrainingParams}. \ed{Future work could further optimize the hyper-parameter selection, or explore training all model parameters.}
We evaluate CROCScore on (1) CROC$^{\text{hum}}$, (2) GenAI-Bench \cite{li2024genaibenchevaluatingimprovingcompositional}, (3) Winoground \cite{thrush_and_ross2022winoground}, and (4) the TIFA benchmark \cite{Hu_2023_ICCV} with original annotations and additional annotations by \citet{cho2024davidsonian}.

\section{Results \& Analysis}
In this section, we analyze 
the metrics' performance on our \cl{datasets}. Further, we discuss the human evaluations and the implications on the usage of auto-generated datasets to evaluate T2I metrics. 

\begin{table}[t]
\centering
\setlength{\tabcolsep}{2.5pt}
\small
\begin{tabular*}{\columnwidth}{@{\extracolsep{\fill}}@{}
      l|ccc|c|cc@{}}
    \toprule
    \multicolumn{1}{c}{} &
    \multicolumn{3}{c}{\textbf{Embedding-based}} &
    \multicolumn{1}{c}{\textbf{Tuned}} &
    \multicolumn{2}{c}{\textbf{VQA-based}} \\
    \textbf{P} & \textbf{Align} & \textbf{BLIP2} & \textbf{CLIP} & \textbf{Pick} & \textbf{BVQA} & \textbf{VQAS} \\
    \midrule
    \multicolumn{7}{l}{\underline{Forward Text-Based}} \\
EV & 0.022 & -0.037 & -0.308 & 0.039 & -0.017 & \textbf{0.552} \\
EP & 0.922 & 0.883 & -0.214 & 0.796 & 0.868 & \textbf{0.968} \\
PV & 0.044 & 0.041 & -0.309 & 0.004 & -0.034 & \textbf{0.442} \\
\midrule
\multicolumn{7}{l}{\underline{Inverse Text-Based}} \\
EV & \textbf{-0.032} & -0.103 & -0.383 & -0.099 & -0.261 & -0.164 \\
EP & \textbf{0.904} & 0.874 & -0.201 & 0.737 & 0.505 & 0.841 \\
PV & \textbf{0.186} & 0.019 & -0.281 & -0.034 & -0.100 & -0.043 \\
\midrule
\multicolumn{7}{l}{\underline{Forward Image-Based}} \\
EV & 0.541 & \textbf{0.585} & 0.402 & 0.564 & 0.438 & 0.580 \\
EP & 0.958 & \textbf{0.959} & 0.662 & 0.906 & 0.872 & 0.946 \\
PV & 0.700 & 0.706 & 0.456 & 0.637 & 0.641 & \textbf{0.720} \\
\midrule
\multicolumn{7}{l}{\underline{Inverse Image-Based}} \\
EV & 0.622 & \textbf{0.657} & 0.304 & 0.529 & 0.554 & 0.642 \\
EP & 0.979 & 0.965 & 0.589 & 0.918 & 0.949 & \textbf{0.984} \\
PV & 0.659 & 0.684 & 0.372 & 0.564 & 0.603 & \textbf{0.733} \\
    \bottomrule
  \end{tabular*}
\caption{Scaled metric accuracy on our supervised dataset. Metrics are abbreviated as \textbf{Align}Score, \textbf{CLIP}Score, \textbf{BLIP2}-ITM, \textbf{Pick}Score, \textbf{BVQA} and \textbf{VQAS}core. The \textbf{P} column denotes the prompt type (abbreviated as: EV = entity variation, EP = entity placement, PV = property variation). The scores are scaled such that 0 indicates random picking, -1 indicates preference for contrast pairs, and 1 indicates correct preference for matching pairs. We bold the highest score for each row if it is higher than 0.}
\label{tab:combined_t2i}
\end{table}
\begin{figure}
    \centering
    \includegraphics[width=\linewidth]{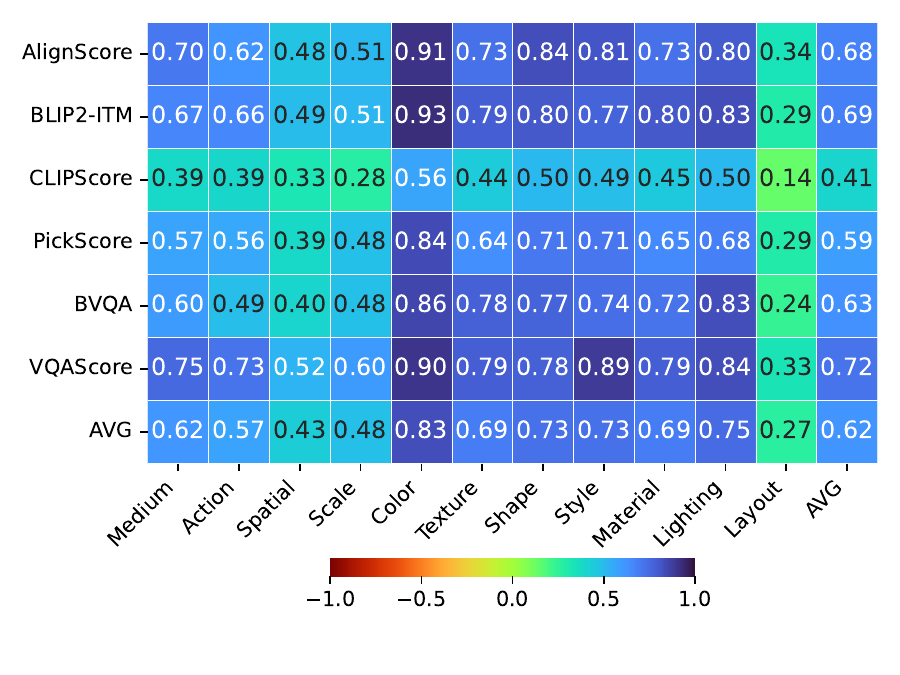}
    \caption{Scaled image-based accuracy per metric on the top-level properties of CROC$^{\text{syn}}$.}
    \label{fig:heatmappseudo}
\end{figure}
\begin{figure*}[htbp]
  \includegraphics[width=\linewidth]{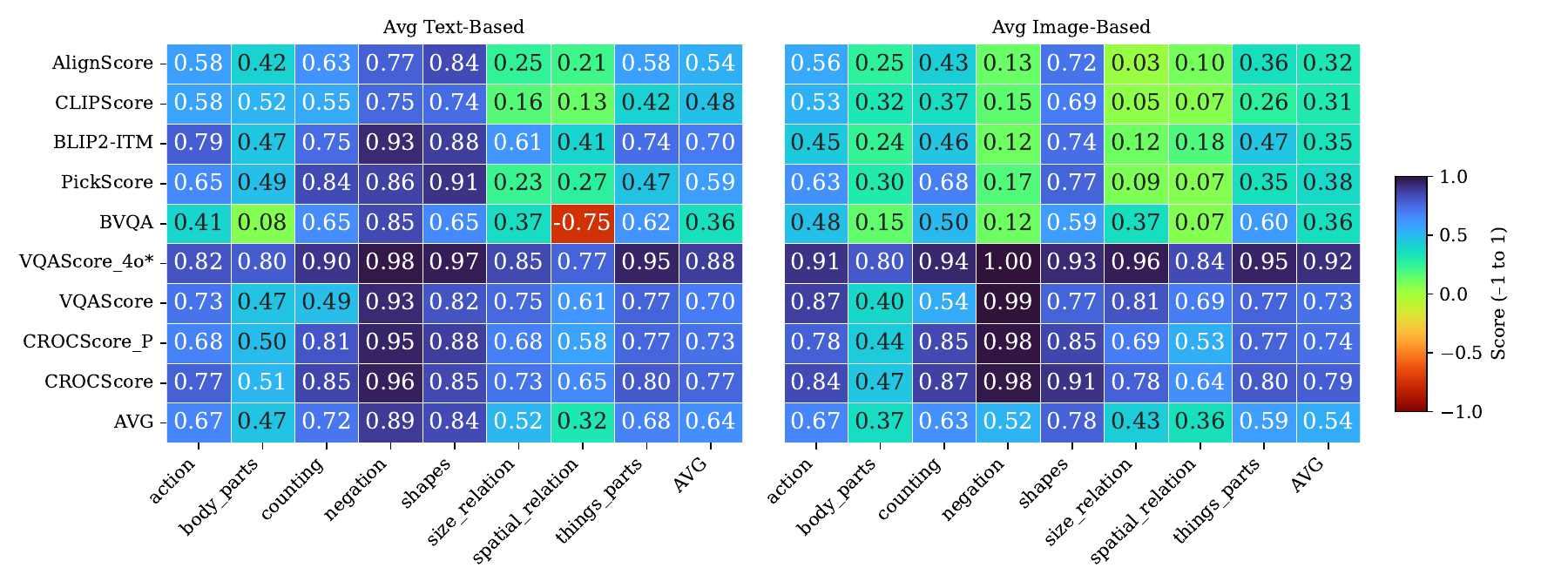}
  \caption{Scaled metric accuracy per category and evaluation direction for CROC$^{\text{hum}}$. For 1, a metric correctly rated all matching pairs higher than the contrast, for 0 it is random and for -1 it rated all contrast pairs higher than the matching pair. The tables show the cell-wise average of the forward and inverse evaluations. *VQAScore\_4o may exhibit mild bias, because a small number of images was created interactively with GPT-4o and GPT-5. Excluding these rows only changes its average by 0.01.}
  \label{fig:supervised_dataset_metric_comp}
\end{figure*}

\subsection{Quantitative Results} \label{quant_res}
\paragraph{CROC$^{\text{syn}}$} Table \ref{tab:combined_t2i} shows the scaled accuracy each metric reaches (averaged across text and image generation models). ``Scaled'' means that we scale the accuracy such that 0 indicates random choice, negative values indicate a preference towards contrast pairs and positive values indicate a preference towards matching pairs.\footnote{As described in \S\ref{evalsetups}, due to the random chance of the evaluation setups, text-based evaluation on CROC$^{\text{syn}}$ is scaled such that 0.83 is mapped to 0, while 0.5 is mapped to 0 for image-based evaluation.} For example, the first table cell indicates that \cll{AlignScore} reaches a scaled accuracy of 0.022 in successfully rating the pair $T_O$-$I_O$ higher than $T_C$-$I_{O}$. Here, the \textbf{P}rompt column indicates the experiment type is \textit{entity variation (EV)}. 
\reb{The metric VQAScore (with a scaled accuracy between $-0.164$ and $0.984$) performs best in 6/12 evaluation setups. Notably, this includes 3/4 property variation directions.
Second-best are the embedding-based metrics BLIP2-ITM (3/12) and  AlignScore (2/12). That means, similar to the findings of \citet{saxon2024who}, embedding-based metrics like AlignScore perform competitive to VQA-based metrics in some settings}. Compared to the VQA-based approaches, \reb{the other types have the benefit of being more resource-efficient (see Table \ref{metricConfig}).} 
 \cll{All \ed{scaled} accuracies for inverse text-based entity variation \ed{in the table} are negative. We hypothesize that the chosen image-generation models were not able to sufficiently generate the entities with changed definitions. The highest accuracy is achieved for entity placement. This is reasonable, due to the clear distinction of environments the entities are placed in. Notably, in text-based evaluation both entity placement and variation exhibit lower inverse than forward accuracies, whereas the relationship reverses under image-based evaluation. This directional asymmetry suggests that the metrics are more robust in judging unexpected text-image pairs when comparing between images than for texts.} \cll{The Kendall \cite{kendall1945ties} correlation between the accuracies for the two text generation models is  \reb{0.872}, i.e., the setup is very stable in that regard.}

\paragraph{Fine-grained analysis - CROC$^{\text{syn}}$}
Figure \ref{fig:heatmappseudo} shows a heatmap of scaled accuracies for the top-level properties of our taxonomy for image-based evaluation, \se{i.e., considering property variation}. 
All values, besides \textit{Layout}, \se{range} between \reb{0.28 and 0.93}. As in Table \ref{tab:combined_t2i}, VQAScore performs best. \textit{Color} is the easiest category, with an average of \reb{0.83}, while \textit{Layout} is the hardest, 
with 0.28. As this is CROC$^{\text{syn}}$, errors may reflect both metric failures and generation failures, although our aggregation mitigates the latter to some extent. However, we can also see that for many cases, some metrics perform better than others. In other words, these can act as an upper bound for performance (unless surpassed in human evaluation). For example, in the \textit{Layout} category,  AlignScore surpasses PickScore by 0.05 and in the \textit{Action} category, VQAScore surpasses AlignScore by 0.11 (Figure \ref{fig:croc-wide} discusses the effect of prompt lengths). These comparisons can guide metric selection for specific use-cases. Future work might explore the usage for metric comparison in targeted domains, such as industry applications. 

\paragraph{Results on CROC$^{\text{hum}}$}
Figure \ref{fig:combined_six_columns} shows an example of six failure cases identified by our setup and Figure \ref{fig:supervised_dataset_metric_comp} gives an overview of the \ed{scaled metric accuracy} on CROC$^{\text{hum}}$. 
We have averaged the results for the forward and inverse setting, because only for \textit{action} (where the contrast includes more unusual actions) and \textit{negation} (where the contrast includes the negation) the inverse direction is expected to perform much differently. 
In this evaluation, we include our own tuned Metric CROCScore and its base metric (Phi4 with contrastive prompting), indicated as CROCScore$_P$. \ed{We do not evaluate and compare CROCScore on CROC$^{\text{syn}}$ because it was trained on it.} 
Additionally, we add VQAScore with GPT-4o backend, which is much more cost and time intensive than the other metrics. Overall, the most difficult categories are \textit{spatial relations} and \textit{body parts}. The simplest categories are \textit{shapes} and text-based \textit{negation}. VQAScore with GPT-4o wins against the open-source metrics by 0.11 and 0.13 points for the text-based and image-based evaluation respectively. Among the open-source metrics, CROCScore has the highest average accuracy, which is 0.04 and 0.05 points above CROCScore\_P. Notably, CROCScore uses only 5.6B parameters compared to VQAScore's CLIP-Flan-T5-XXL with 11.3B parameters, i.e., it is also more resource-efficient. An interesting case is BVQA for spatial relations, which shows a negative scaled accuracy, indicating that the metric has learned to inverse spatial relations. Only the VQAScore and CROCScore metrics can adequately handle the tricky inverse image-based negation case, where the metric needs to decide that something is not in the picture (all other metrics are lower than 0.17 for image-based negation). Interestingly, AlignScore does not perform as well as on the CROC$^{\text{syn}}$ dataset. \ed{The underlying Align model \cite{pmlr-v139-jia21b} was trained on noisy, contrastive data. Perhaps, this makes the model more in-domain on CROC$^{\text{syn}}$, while fine image details, like body parts, are less considered. \reb{Likewise, BLIP2-ITM only performs competitive on the text-based setup.}} 
For VQAScore\se{,} \cll{35\% of the} setups have a value below 0.6, that is, (because of scaling) at least one out of 5 samples failed.  
For CLIPScore\se{,} it is \cll{81.3\% of the} setups. The GPT-4o VQAScore is below 0.6 in none of the cases.  
Overall, we can see (uniform colors) that \ed{the weaker block of the} 
upper five \ed{and the block of the lower four} metrics (VQAScore and CROCScore) have similar failure rates. \ed{In Figure \ref{fig:metriccorrs}, we underline this finding with pairwise metric correlations. For image-based evaluation, all metrics in the upper block have averages below or equal to 0.4 scaled accuracy, while the block of VQAScore and CROCScore has at least 0.7. For text-based evaluation, this gap is much smaller: BLIP2-ITM reaches an average of 0.7. In other words, image-based evaluation is more difficult for embedding-based metrics. On the other hand, VQA-based metrics achieve about 0.03 points less for text-based evaluation. For action and shape, the embedding-based metrics even outperform the VQA-based metrics in text-based evaluation.} 
The fine-tuning of CROCScore improves the \ed{untuned CROCScore$_p$} in 14 of 16 categories, 
i.e., it increases the robustness of the metric (for example, action is increased by 0.09 and counting by 0.04). Further, CROC$^{\text{syn}}$ contains images generated by models with performance-runtime tradeoff. Future work could further explore the use of stronger T2I models to tune advanced metrics.

\begin{table}[t]
\centering
\small
\setlength{\tabcolsep}{2.5pt}
\begin{tabular}{l|c|c}
    \toprule
    \textbf{Metric} & \textbf{Kendall $\tau_B$} & \textbf{Pairwise Acc.} \\
    & (Basic/Adv/Overall) & (Basic/Adv/Overall) \\
    \midrule
    Align & 0.072 / 0.113 / 0.131 & 0.481 / 0.503 / 0.514 \\
    BLIP2 & 0.081 / 0.485 / 0.147 &  0.141 / 0.522 / 0.516 \\
    VQAS & 0.403 / 0.310 / 0.398 & 0.637 / 0.596 / 0.641 \\
    CROC\_P & 0.443 / 0.386 / 0.452 &  0.623 / 0.624 / 0.650 \\
    CROC & \textbf{0.446} / \textbf{0.401} / \textbf{0.454}  & \textbf{0.649} / \textbf{0.631} / \textbf{0.660} \\
    \bottomrule
\end{tabular}
\caption{Results of \reb{\textbf{Align}Score}, \textbf{BLIP2}-ITM, \textbf{VQAS}core, \textbf{CROC}Score$_\text{plain}$ and \textbf{CROC}Score on GenAi-Bench \citep{li2024genaibenchevaluatingimprovingcompositional}, where plain refers to the metric without fine-tuning. Basic, Advanced and Overall are categories in GenAi-Bench. The reported measures are Kendall correlation and Pairwise Accuracy \cite{deutsch-etal-2023-ties}.} 
\label{tab:kendall_pairwise}
\end{table}
    
\begin{table}[t]
\centering
\small
\setlength{\tabcolsep}{2.5pt}
\begin{tabular}{l|c|c|c}
    \toprule
    \textbf{Metric} & Text & Image & Group  \\
    \midrule
    Align & 0.353 & 0.105 & 0.0825 \\
    BLIP2 & 0.428 & 0.223 & 0.183 \\
    VQAS & 0.605 & \textbf{0.575} & 0.458 \\
    CROC\_P & 0.587 & 0.522 & 0.428 \\
    CROC & \textbf{0.615} & 0.558 & \textbf{0.468} \\
    \bottomrule
\end{tabular}
\caption{Results of \textbf{AlignScore}, \textbf{BLIP2}-ITM, \textbf{VQAScore}, \textbf{CROCScore}$_\text{plain}$, and \textbf{CROCScore} on Winoground \cite{thrush_and_ross2022winoground}. Winoground also contains contrastive text-image pairs; the reported values are accuracies for distinguishing matched from mismatched pairs in both \textit{Text} directions, both \textit{Image} directions or all four directions (\textit{Group}).}
\label{tab:winoground}
\end{table}

\begin{table}[t]
\centering
\small
\setlength{\tabcolsep}{2.5pt}
\begin{tabular}{l|c|c}
    \toprule
    \textbf{Metric} & \textbf{Original (K/P)} & \textbf{DSG (K/P)} \\
    \midrule
    Align   & 0.299/0.408 & 0.276/0.393 \\
    BLIP2 & 0.294/0.409 & 0.233/0.357 \\
    VQAS    & 0.532/0.657 & 0.527/0.665 \\
    CROC\_P & 0.549/0.645 & \textbf{0.538}/0.641 \\
    CROC    & \textbf{0.550}/\textbf{0.680} & 0.532/\textbf{0.675} \\
    \bottomrule
\end{tabular}
\caption{Results of \textbf{Align}Score, \textbf{BLIP2}-ITM, \textbf{VQAS}core, \textbf{CROC}Score$_\text{plain}$ and \textbf{CROC}Score on TIFA \cite{Hu_2023_ICCV}. We show the \textbf{K}endall and \textbf{P}earson correlations. The \textit{Original} column shows the correlation to scores in the original TIFA paper by \citet{Hu_2023_ICCV}. The \textit{DSG} column shows the correlation to the re-annotated data by \citet{cho2024davidsonian}.} 
\label{tab:tifa}
\end{table}

\begin{table}[t]
\centering
\small
\setlength{\tabcolsep}{2.5pt}
\begin{tabular}{l|c|c|c|c|c}
    \toprule
    \textbf{Metric} & F.\ T.\ & I.\ T.\ & F.\ I.\ & I.\ I.\ & Avg \\
    \midrule
    Human$_p$  & \textbf{0.659} & \textbf{0.522} & 0.590 & 0.526 &  \textbf{0.574} \\
    Align  & 0.145 & 0.120 & 0.707 & 0.695 & 0.417\\
    BLIP2      & 0.036 & 0.048 & 0.647 & 0.727 & 0.364 \\
    VQAScore   & 0.421 & -0.039 & \textbf{0.711} & \textbf{0.755} & 0.462 \\
    \midrule
    Human$_i$  & \textbf{0.832} & \textbf{0.684} & \textbf{0.768} &\textbf{ 0.705} & \textbf{0.738} \\
    Human$_p$  & 0.600 & 0.411 & 0.495 & 0.495 & 0.500 \\
    Align & -0.027 & 0.053 & 0.663 & 0.684 & 0.334 \\
    BLIP2      & -0.065 & 0.053 & 0.558  & 0.642  & 0.355  \\
    VQAScore   & 0.305 & -0.091 & 0.705 & 0.579 & 0.380 \\    
    \bottomrule
\end{tabular}
\caption{\textbf{Human}, \textbf{Align}Score, \textbf{BLIP2}-ITM and \textbf{VQAScore} scaled accuracy on CROC$^{\text{syn}}$ in the directions \textbf{F}orward \textbf{T}ext-Based, \textbf{I}nverse \textbf{T}ext-Based, \textbf{F}orward \textbf{I}mage-Based and \textbf{I}nverse \textbf{I}mage-Based. The upper part shows scores on 2000 text-image pairs annotated by Prolific annotators (Human$_p$). The lower part shows scores on 380 text-image pairs that were additionally annotated by in-house annotators (Human$_i$).}
\label{tab:human_eval}
\end{table}

\paragraph{CROCScore on further benchmarks}
\reb{To verify the strong performance of CROCScore, we evaluate it on (1) GenAI-Bench (Table \ref{tab:kendall_pairwise}), (2) Winoground (Table \ref{tab:winoground}), and (3) the TIFA benchmark (Table \ref{tab:tifa}). On most of these benchmarks, CROCScore achieves higher scores than the other open-source metrics. Only for the Kendall score of the DSG TIFA annotations and for \textit{Image} on Winoground, other metrics win. This highlights the value of CROC$^\text{syn}$ for tuning new metrics.}

\paragraph{Comparison: CROC$^{\text{syn}}$ vs.\ CROC$^{\text{hum}}$}
\ed{To evaluate 
how correlated CROC$^{\text{syn}}$ is with CROC$^{\text{hum}}$, we first select property groups in \reb{CROC$^{\text{syn}}$} that are similar to the properties in CROC$^{\text{hum}}$: (1) action, (2) spatial, (3) size, (4) shape and (5) negation. For each combination of these five groups and the 6 metrics evaluated on CROC$^{\text{syn}}$, we compute the average accuracy. Then, we compare these accuracies, i.e., these five rankings, with the corresponding categories in Figure \ref{fig:supervised_dataset_metric_comp}. To do so, we compute the Kendall correlation of the flattened accuracy tables. For text-based evaluation, this yields a Kendall agreement of 0.40 (p<0.003) and for image-based it yields a Kendall agreement of 0.52 (p<0.0001). In other words, even though (1) the categories are not exactly the same, (2) the prompts of CROC$^{\text{syn}}$ are more detailed and (3) these categories can be difficult to generate, the metric rankings per category are similar. 
The text-based agreement \se{might be} lower because of this different complexity in dataset texts.} 

\subsection{Human evaluation}
\label{sub_sec:human_eval}
Table \ref{tab:human_eval} shows the results of our human evaluation on CROC$^{\text{syn}}$. On a set of 2,000 text-image pairs, Prolific annotators achieve higher scaled accuracy than VQAScore on text-based items but lower scaled accuracy on image-based items. Krippendorff's alpha per batch ranges from $-0.05$ to $0.656$ (median $0.404$) across 26 batches, indicating that our attention checks only partially mitigated variability. To obtain an additional higher-quality reference, we label a subset of 380 image pairs with three in-house annotators. They outperform all automatic metrics in every category. Their agreement is  $\alpha=0.604$. TIFA ($0.67$) and GenAI-Bench ($0.72$) report comparable agreements as substantial. The human in-house performance indicates that CROC$^{\text{syn}}$ can successfully reveal metric shortcomings, because metrics have not reached the upper boundary.

In addition, we measure the disagreement for each property on the in-house sample as the \emph{mean absolute difference} between per-sample annotator scores. The highest disagreement occurs for the properties \textit{right-of} and \textit{close proximity}. One sample, e.g., swaps \textit{the position of musician in a concert}. We verify that images are generated as intended. However, the prompt features 93 words, so the disagreement might be caused by the contrast either being overlooked or being rated differently in relation to the other, correctly generated, parts of the prompt. 

We also perform a human \ed{evaluation} on \textbf{CROC$^{\text{hum}}$}. Here, the three annotators achieve a Krippendorff's Alpha of 0.957 and an \ed{average scaled accuracy (from $-1$ to $1$) of 0.949}, highlighting the quality of the dataset. \fnl{Notably, this human accuracy is largely better than the metric performance in Figure \ref{fig:supervised_dataset_metric_comp}, i.e., about 0.1 points better than VQAScore with GPT-4o.}

\section{Conclusion}
\label{sec:conclusion}
\cl{We introduce \vrobtitle, a meta-evaluation framework for T2I metrics that enables fine-grained robustness tests based on contrastive text-image pairs. We use this framework to generate CROC$^{\text{syn}}$, a novel large-scale dataset that we use to evaluate existing metrics and to train our new metric CROCScore. The evaluation shows that our approach can be used to successfully compare metric performance. Further, we introduce CROC$^{\text{hum}}$, a contrastive human-supervised dataset of challenge categories that allows 
a fine-grained analysis of metric failure cases. \fnl{Both datasets show that all tested metrics still exhibit blindspots and do not reach human performance. This is especially true for body parts and spatial relations, although VQAScore with a GPT-4o backend is already much more robust.}
Our metric CROCScore achieves state-of-the-art results among the tested open-source metrics on CROC$^{\text{hum}}$ dataset and GenAI-bench \cite{li2024genaibenchevaluatingimprovingcompositional}. \ed{Further, it provides a cost advantage over GPT-4o.} This opens an interesting future research path of improving VQA-based T2I metrics by tuning on fine-grained generated contrastive datasets.} 

\section*{Acknowledgements}
The NLLG group gratefully acknowledges support from the Federal Ministry of Education and Research (BMBF) via the research grant ``Metrics4NLG'' and the German Research Foundation (DFG) via the Heisenberg Grant EG 375/5-1. Further, we thank the annotators of our human evaluations. Also, we thank the action editor and the reviewers for their constructive feedback that has helped us to improve our work. The authors also acknowledge support by the state of Baden-Württemberg through bwHPC and the German Research Foundation (DFG) through grant INST 35/1597-1 FUGG.

\bibliography{tacl2021}
\bibliographystyle{acl_natbib}

\appendix

\section{Taxonomy Properties}
\label{app:taxProperties}
Here, we give an overview of all taxonomy properties and example topics (with 2 example entities each):
{\small
\begin{itemize}[label={},leftmargin=1em,itemsep=0.1em,topsep=0.1em,parsep=0.4pt,partopsep=0pt]
\item \textbf{Example Topics}: 
  \textit{Nature:} Landscapes [Mountain, River], Flora [Tree, Flower], Fauna [Deer, Eagle], Weather Phenomena [Lightning Bolt, Snowflake], Underwater [Sea Turtle, Coral]; 
  \textit{People:} Portraits [Adult Human, Child], Groups [Friends, Crowd], ...; 
  \textit{Animals:} Wild Animals [Lion, Elephant], Domestic Animals [Dog, Cat], ...; 
  \textit{Architecture:} Residential Buildings [House, Cottage], Commercial Buildings [Skyscraper, Shopping Mall], ...
\end{itemize}

\textbf{Property - \textit{Medium}}: 
\textit{Photography}, \textit{Illustration}, \textit{3D Rendering}, \textit{Anime}, \textit{Mixed Media}, \textit{Painting}

\textbf{Property - \textit{Relation}}
\begin{itemize}[label={},leftmargin=1em,itemsep=0.1em,topsep=0.1em,parsep=0.4pt,partopsep=0pt]
  \item \textbf{Action}: 
    \textit{Gesture:} Pointing, Waving, Facial Expression, Nodding; 
    \textit{Full‑Body Movement:} Running, Dancing, Jumping, Swimming
  \item \textbf{Spatial}: 
    \textit{Foreground/Background:} Foreground Emphasis, Midground Placement, Background Silhouette; 
    \textit{Proximity/Overlap:} Close Proximity, Overlapping Forms, Left‑of, Right‑of, Above, Below, Inside
  \item \textbf{Scale}: 
    \textit{Exaggerated:} Giant Figures, Miniature Objects, Distorted Perspective; 
    \textit{Realistic Scale:} Life‑Size Representation, Proportional Figures, Consistent Depth
\end{itemize}

\textbf{Property - \textit{Attribute}}
\begin{itemize}[label={},leftmargin=1em,itemsep=0.1em,topsep=0.1em,parsep=0.4pt,partopsep=0pt]
  \item \textbf{Color}: 
    \textit{Monochrome}, \textit{Vibrant}, \textit{Red}, \textit{Blue}, \textit{Green}, \textit{Yellow}, \textit{Purple}, \textit{Orange}, \textit{Pink}, \textit{Brown}, \textit{Black}, \textit{White}
  \item \textbf{Texture}: 
    \textit{Smooth}, \textit{Rough}, \textit{Reflective}
  \item \textbf{Shape}: 
    \textit{Geometric}, \textit{Organic}
  \item \textbf{Style}: 
    \textit{Realistic}, \textit{Impressionistic}, \textit{Minimalist}
  \item \textbf{Material}: 
    \textit{Metallic}, \textit{Wooden}, \textit{Fabric}, \textit{Plastic}, \textit{Glass}, \textit{Stone}, \textit{Paper}
  \item \textbf{Lighting}: 
    \textit{Natural Light}, \textit{Artificial Light}, \textit{High Contrast}
  \item \textbf{Layout}: 
    \textit{Centered}, \textit{Rule of Thirds}, \textit{Asymmetrical}
\end{itemize}
} 
\begin{table*}[ht]
  \centering
  \small
  \begin{tabular}{@{} llp{7.2cm}l @{}}
    \toprule
    \textbf{Metric} & \textbf{Type} & \textbf{Version and Model(s)} & \textbf{Runtime} \\
    \midrule
    CLIPScore & Embed 
      & \href{https://github.com/Lightning-AI/torchmetrics/releases/tag/v1.6.2}{torchmetrics\_1.6.2}; \href{https://huggingface.co/openai/clip-vit-large-patch14}{\texttt{clip-vit-large-patch14}}
      & ca.\ 275 Sec.\\
    BLIP2-ITM & Embed
      & \href{https://github.com/linzhiqiu/t2v_metrics}{VQAScore (Implementation) from 3.25}; \href{https://huggingface.co/Salesforce/blip2-itm-vit-g}{\texttt{blip2-itm-vit-g}}
      & ca.\ 112 Sec.\\
    ALIGNScore & Embed 
      & \href{https://github.com/michaelsaxon/T2IScoreScore}{T2IScoreScore 1.25};\href{https://huggingface.co/kakaobrain/align-base}{\texttt{align-base}}
      & ca.\ 70 Sec.\\
    PickScore & Tuned 
      & \href{https://huggingface.co/yuvalkirstain/PickScore\_v1}{\texttt{PickScore\_v1}}
      & ca.\ 61 Sec.\\
    VQAScore & VQA
      & \href{https://github.com/linzhiqiu/t2v\_metrics}{VQAScore from 03.25};\href{https://huggingface.co/zhiqiulin/clip-flant5-xxl}{\texttt{clip-flant5-xxl}}
      & ca.\ 40 Min. (no batching) \\
    VQAScore\_4o & VQA
      & \href{https://github.com/linzhiqiu/t2v\_metrics}{VQAScore from 03.25};\href{https://openai.com/index/hello-gpt-4o/}{\texttt{GPT-4o (04.25)}}
      & ca.\ 55 Min. \\
    BVQA & VQA
      &  \href{https://github.com/Karine-Huang/T2I-CompBench}{T2I-CompBench from 03.25}
      & ca.\ 22.7 Min.\\
    CROCScore & Tuned VQA 
      & \href{https://huggingface.co/microsoft/Phi-4-multimodal-instruct}{microsoft/Phi-4-multimodal-instruct}, batch size=4
      & ca.\ 11 Min.\\
    \bottomrule
  \end{tabular}
  \caption{Overview of the metrics we evaluate, with a brief description, key configuration details, and their runtime on 1\,000 “body parts” samples from CROC$^{hum}$. Metrics that directly return the quality score are  faster than VQA based metrics, because they do not require auto-regressive generation.}
  \label{metricConfig}
\end{table*}
\section{Inverse Equations}
Here we show the equation that we apply for the evaluation of inverse text-base samples. $M$ is a metric, $T_O$ and $I_O$ are the original text and image, $T_C$ and $I_C$ are the contrast text and image. $i$ and $j$ are indices for one of multiple images:
\label{inverseEquations}
\begin{align}
j^* = \underset{i=1,\ldots,n}{\operatorname{argmax}}\, \cl{M}(\cll{T_C, I^i_C}),\nonumber\\
\cl{M}(\cll{T_C, I^{j^*}}) > \cl{M}(\cll{T_O, I^{j^*}})
\end{align}

This is the respective equation for inverse image-based evaluation:
\begin{equation}
\underset{i=1,\ldots,n}{\operatorname{max}}\, \cl{M}(\cll{T_C, I^i_C}) > \underset{i=1,\ldots,n}{\operatorname{max}}\, \cl{M}(\cll{T_C, I^i_O})
\end{equation}

\section{Detailed examples for generation and evaluation}
\label{DetailedExamples}
\cll{In the following, we present one example of data construction and evaluation with our pseudo-labeled generation process (for CROC$^\text{syn}$) and one example with our human-supervised generation process (for CROC$^\text{hum}$).}

\begin{figure*}[htbp]
  \centering
  \begin{minipage}[t]{0.18\textwidth}
    {\small\textbf{Original 1 ($I_O^1$)}}\\
    \includegraphics[width=\linewidth]{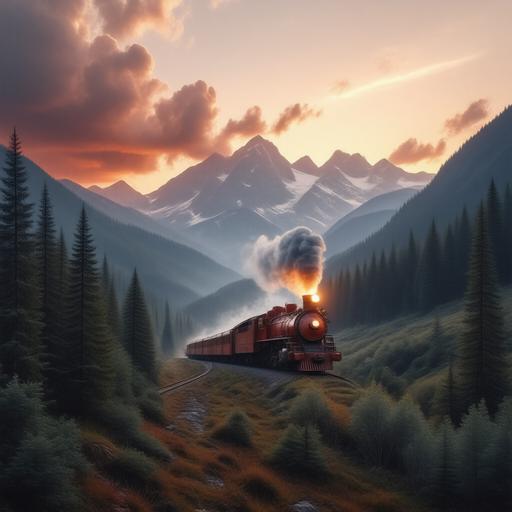}\\[-1.7mm]
    {\small\tikzmark{boxstart}$M(T_O,I^1_O)=13.9$}\\[-0.6mm]
    {\small $M(T_C,I^1_O)=14.3$}\\
    {\small\textbf{Contrast 1 ($I_C^1$)}}\\
    \includegraphics[width=\linewidth]{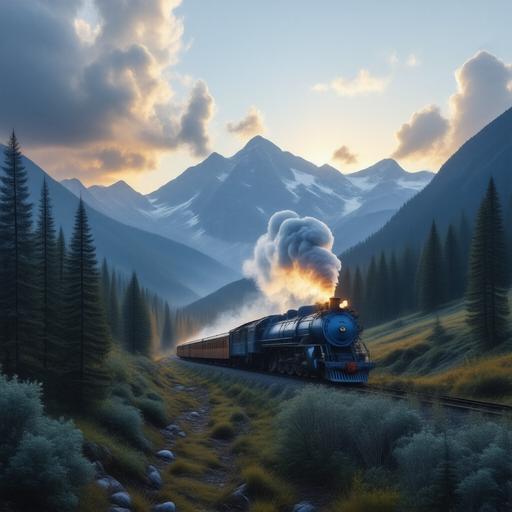}\\[-1.7mm]
    {\small $M(T_O,I^1_C)=14.9$}\\[-0.5mm]
    {\small $M(T_C,I^1_C)=12.7$}
  \end{minipage}\hfill
  \begin{minipage}[t]{0.18\textwidth}
    {\small\textbf{Original 2 ($I_O^2$)}}\\
    \includegraphics[width=\linewidth]{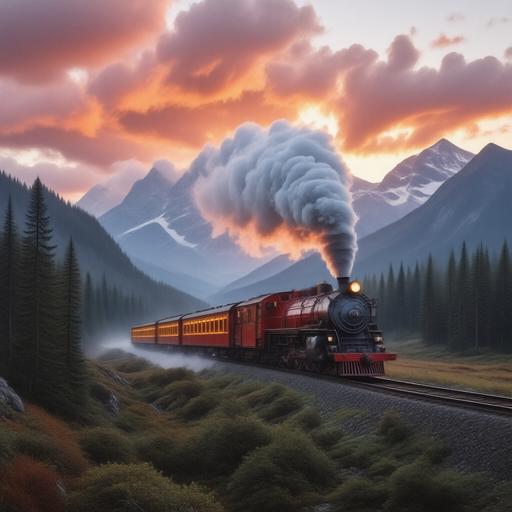}\\[-1.7mm]
    {\small\tikzmark{redstart}$M(T_O,I^2_O)=14.4$}\\[-0.6mm]
    {\small $M(T_C,I^2_O)=14.7$\tikzmark{redend}}\\
    {\small\textbf{Contrast  2 ($I_C^2$)}}\\
    \includegraphics[width=\linewidth]{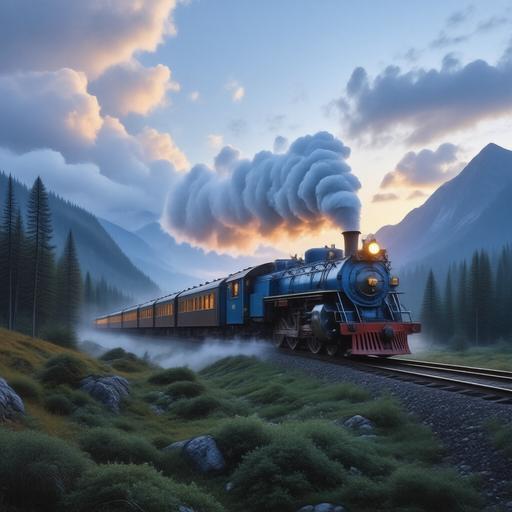}\\[-1.7mm]
    {\small $M(T_O,I^2_C)=15.2$}\\[-0.5mm]
    {\small $M(T_C,I^2_C)=13.0$}
  \end{minipage}\hfill
  \begin{minipage}[t]{0.18\textwidth}
    {\small\textbf{Original 3 ($I_O^3$)}}\\
    \includegraphics[width=\linewidth]{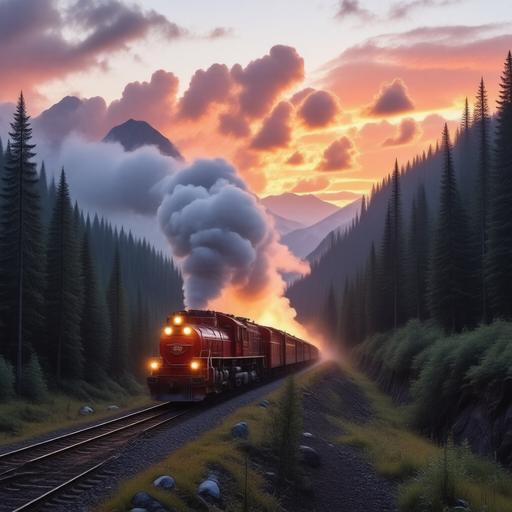}\\[-1.7mm]
    {\small $M(T_O,I^3_O)=14.1$}\\[-0.6mm]
    {\small $M(T_C,I^3_O)=14.6$}\\
    {\small\textbf{Contrast 3 ($I_C^3$)}}\\
    \includegraphics[width=\linewidth]{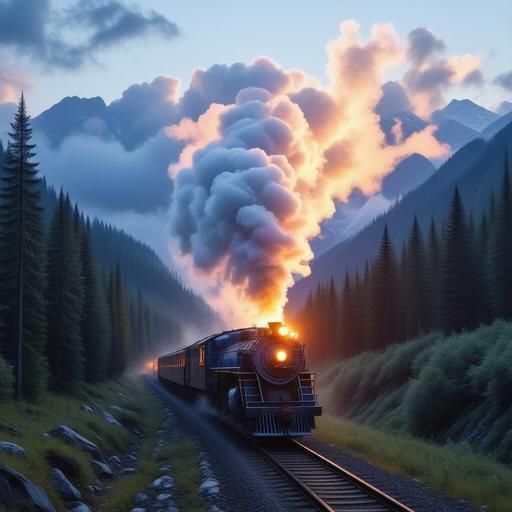}\\[-1.7mm]
    {\small $M(T_O,I^3_C)=15.7$}\\[-0.5mm]
    {\small $M(T_C,I^3_C)=13.9$}
  \end{minipage}\hfill
  \begin{minipage}[t]{0.18\textwidth}
    {\small\textbf{Original 4 ($I_O^4$)}}\\
    \includegraphics[width=\linewidth]{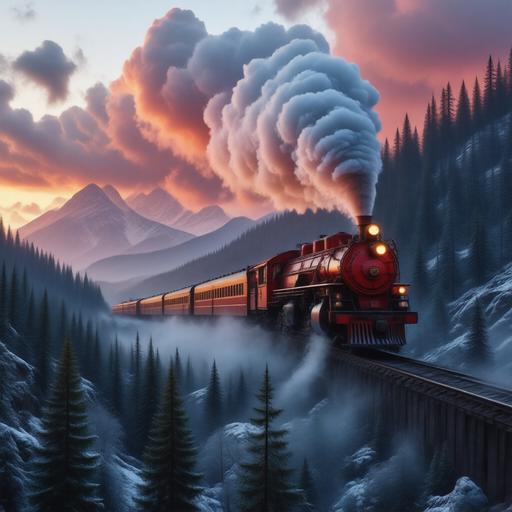}\\[-1.7mm]
    {\small $M(T_O,I^4_O)=13.7$}\\[-0.6mm]
    {\small $M(T_C,I^4_O)=14.6$}\\
    {\small\textbf{Contrast 4 ($I_C^4$)}}\\
    \includegraphics[width=\linewidth]{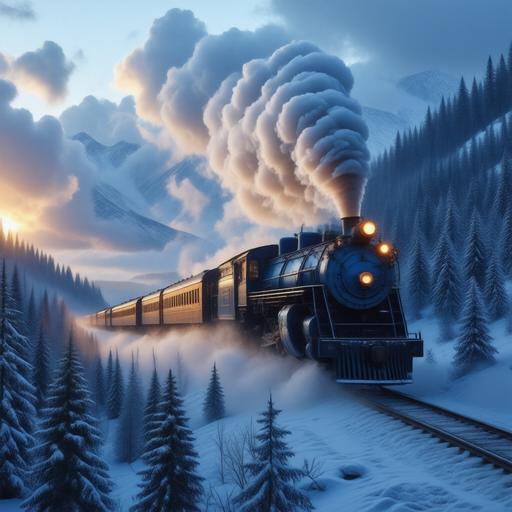}\\[-1.7mm]
    {\small $M(T_O,I^4_C)=13.9$}\\[-0.5mm]
    {\small $M(T_C,I^4_C)=11.7$}
  \end{minipage}\hfill
  \begin{minipage}[t]{0.18\textwidth}
    {\small\textbf{Original 5 ($I_O^5$)}}\\
    \includegraphics[width=\linewidth]{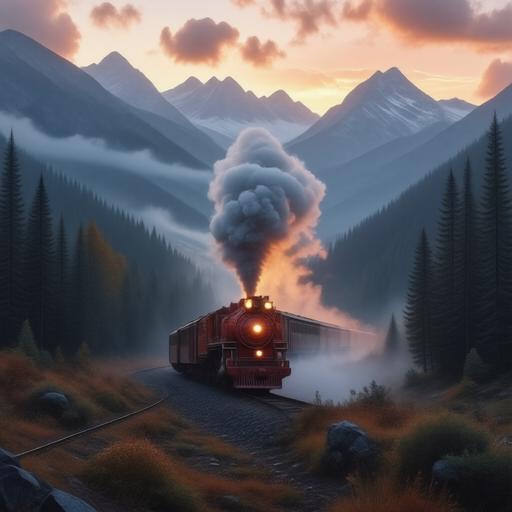}\\[-1.7mm]
    {\small $M(T_O,I^5_O)=12.8$\tikzmark{boxend}}\\[-0.6mm]
    {\small $M(T_C,I^5_O)=13.7$}\\
    {\small\textbf{Contrast 5 ($I_C^5$)}}\\
    \includegraphics[width=\linewidth]{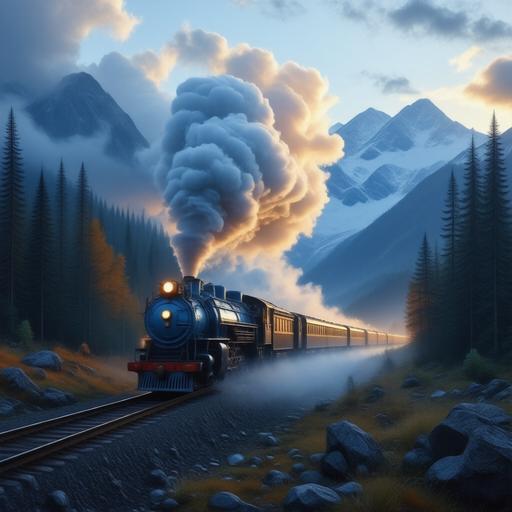}\\[-1.7mm]
    {\small $M(T_O,I^5_C)=14.6$}\\[-0.5mm]
    {\small $M(T_C,I^5_C)=12.5$}
  \end{minipage}
  \begin{tikzpicture}[remember picture,overlay]
    \draw[green,line width=0.2mm]
      ([yshift=8.5pt]pic cs:boxstart) rectangle
      ([yshift=-3.1pt]pic cs:boxend);
    \draw[red,line width=0.2mm]
      ([yshift=8.5pt]pic cs:redstart) rectangle
      ([yshift=-3pt]pic cs:redend);
  \end{tikzpicture}
  \caption{Generated original and contrast images for property variation with subject \textit{Transportation} and property \textit{Red}. $T_O$: \textit{A majestic red steam locomotive chugging through a mountain valley[...]}. $T_C$: \textit{A majestic \cll{blue} steam locomotive chugging through a mountain valley[...]}. Further, we display the metric scores for AlignScore for all combinations. In \textbf{text-based forward} evaluation, we first find the highest value for $T_O$-$I^i_O$ pairs (\green{green} box) and then compare it to the respective $T_C$-$I^i_C$ pair (\red{red} box).}
  \label{fig:locomotiveSample}
\end{figure*}

\paragraph{Pseudo-Labeled Generation - Property Variation}
\begin{enumerate}
    \item \textbf{Property and subject selection} This is an example for \textit{property variation, where w}e select the property ``red'' and the subject ``Transportation''.
    \item \textbf{Prompt generation} In this example, we generate an image with stable diffusion. Hence, we load the stable diffusion guide. Then we fill the \textit{property variation} prompt template from Appendix \ref{promptTemplates} with our data and pass it to an LLM, here the Deepseek model, to generate 5 outputs. One valid output is the following JSON:\\
    \{
    \quad ``prompt \cll{($T_O$)}'': ``A majestic red steam locomotive chugging through a mountain valley[...]'',\\
    \quad ``contrast\_prompt \cll{($T_C$)}'': ``A majestic blue steam locomotive chugging through a mountain valley[...]''
    \}
    \item \textbf{Image generation} Then, we generate 5 images from the extracted prompt and contrast prompt each (see Figure \ref{fig:locomotiveSample}).
    \item \textbf{Metric computation} Next, we compute the metric scores for all Text-Image combinations.
    \item \textbf{Metric evaluation} Finally, we evaluate the metric(s) based on the score. For example, for \textbf{forward text-based} evaluation we first \cll{select the highest $M(T_O,I^{i}_O)$ (green box in Figure \ref{fig:locomotiveSample} and then compare it to the respective $M(T_O,I^{i}_C)$ score (red box). In the example $M(T_O,I^{2}_O)=14.4$ is smaller than $M(T_C,I^{2}_O)=14.7$, therefore the metric did not pass the test case.}\footnote{Metric scores are not always scaled between 0 and 1.} 
\end{enumerate}

\begin{figure*}[htbp]
  \centering
  \noindent
  \begin{minipage}[t]{0.45\textwidth}
    \centering
    \textbf{Original Outputs}\\
    \begin{minipage}[t]{0.32\linewidth}
      \centering
      \includegraphics[width=\linewidth]{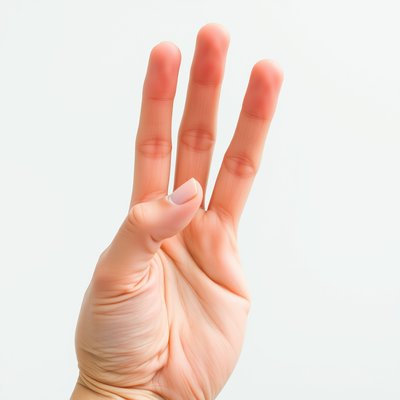}\\[1mm]
      \small Invalid image
    \end{minipage}\hfill
    \begin{minipage}[t]{0.32\linewidth}
      \centering
      \includegraphics[width=\linewidth]{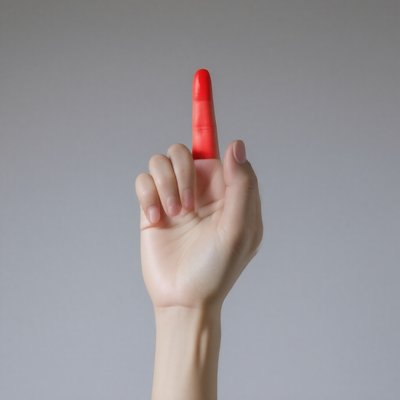}\\[1mm]
      \small $M(T_O,I^1_O)=18.3$
    \end{minipage}\hfill
    \begin{minipage}[t]{0.32\linewidth}
      \centering
      \includegraphics[width=\linewidth]{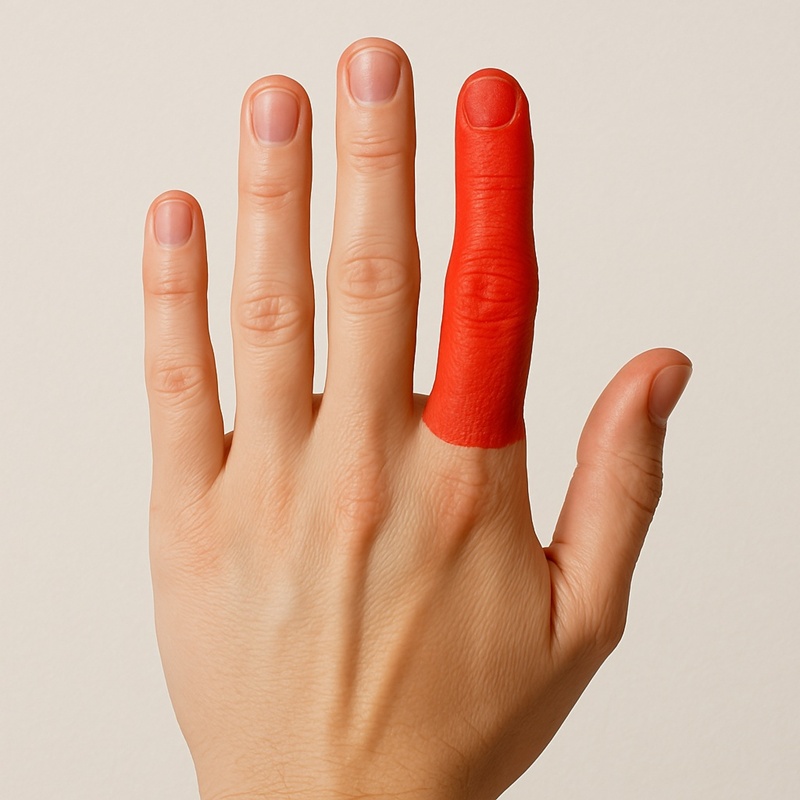}\\[1mm]
      \small $M(T_O,I^2_O)=16.8$
    \end{minipage}
  \end{minipage}%
  \quad\vrule width 0.2pt\quad
  \begin{minipage}[t]{0.45\textwidth}
    \centering
    \textbf{Contrast Outputs}\\
    \begin{minipage}[t]{0.32\linewidth}
      \centering
      \includegraphics[width=\linewidth]{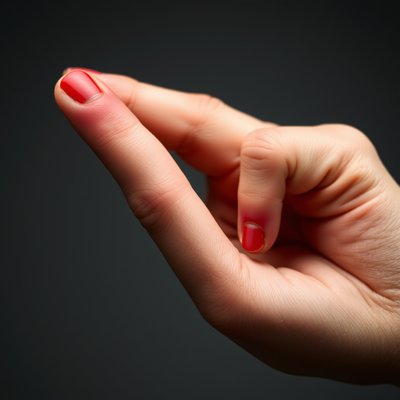}\\[1mm]
      \small Invalid image
    \end{minipage}\hfill
    \begin{minipage}[t]{0.32\linewidth}
      \centering
      \includegraphics[width=\linewidth]{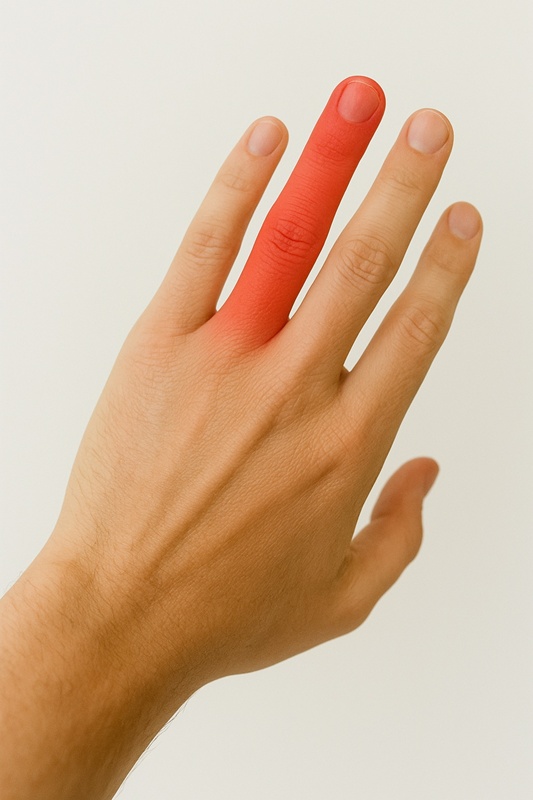}\\[1mm]
      \small $M(T_O,I^1_C)=17.2$
    \end{minipage}\hfill
    \begin{minipage}[t]{0.32\linewidth}
      \centering
      \includegraphics[width=\linewidth]{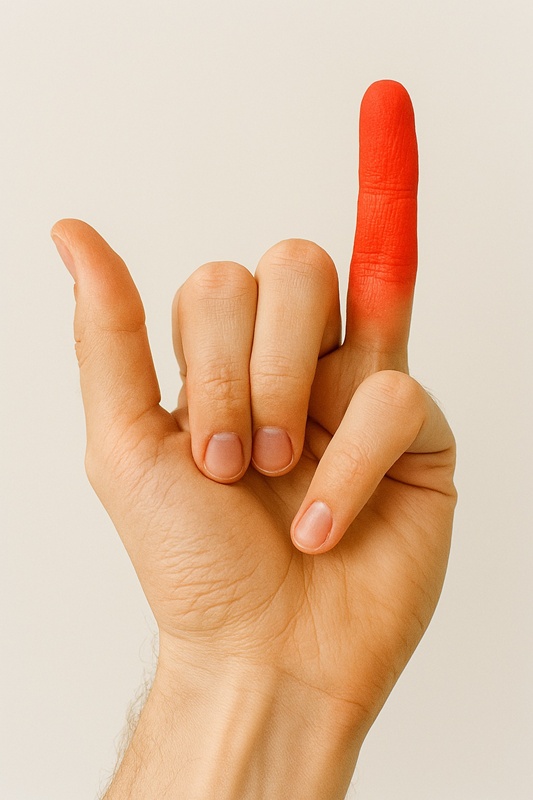}
      \small $M(T_O,I^2_C)=18.2$
    \end{minipage}
  \end{minipage}

  \caption{Example images of the \textit{body parts} test case \cll{(per-default GPT-4o sometimes generates non-square image dimensions)}. $T_O$: \textit{A hand with only its \textbf{index finger} colored red}. $T_C$: \textit{A hand with only its \textbf{ring finger} colored red}. Further, we display the AlignScore scores necessary for forward image‐based evaluation \cll{where we compare each matching score of the original images (left side) with each contrast score of the contrast outputs (right side). That means, 18.3 is (1) higher than 17.2 and (2) higher than 18.2, but 16.8 is (3) lower than 17.2 and (4) lower than 18.2. Therefore the overall accuracy is $\frac{2}{4}$.} }
  \label{fig:body_parts_split}
\end{figure*}

\paragraph{Human-Supervised Generation Example}
\begin{enumerate}
    \item \textbf{Supervised prompt construction} In this example, we choose a prompt and contrast prompt of the category \textit{body parts} that was created through interactive querying of GPT-4o. \textbf{Prompt ($T_O$):} ``A hand with only its index finger colored red.'' \textbf{Contrast ($T_C$)}: ``A hand with only its ring finger colored red.'' 
    \item \textbf{Image generation} Next, we generate 100 images for each prompt. Figure \ref{fig:body_parts_split} shows examplary generations.
    \item \textbf{Supervised image filtering} Then, we manually remove all images that are not matching the prompts. In Figure \ref{fig:body_parts_split}, these are titled ``Invalid image''.
    \item \textbf{Metric computation} Next, we compute the metric scores.
    \item \textbf{Metric evaluation} Here, we demonstrate image-based evaluation. \cll{To calculate the accuracy, we first compare all $M(T_O,I^i_O)$ with all $M(T_O,I^i_C)$. That means, for the four valid images in Figure \ref{fig:body_parts_split} we compare each score of the original outputs (left) with each score of the contrast outputs (right): (1) $18.3>17.2$, (2) $18.3>18.2$, (3) $16.8>17.2$ and (4) $16.8>18.2$. Because two of these four conditions are true, the final score is $\frac{2}{4}$.}
\end{enumerate}

\begin{table}[t]
  \centering
  \begin{tabular}{lrrrr}
    \hline
           & \textbf{AlignScore} & \textbf{CLIPScore} & \textbf{PickScore} \\
    \hline
    \textbf{F.\ T.} & 0.648  & 0.021  & 0.431 \\
    \textbf{I.\ T.} & -0.277 & -0.048 & -0.308 \\
    \hline
  \end{tabular}
  \caption{\textbf{Initial Experiment:} Scaled accuracy on contrastive text-image pairs generated from T2ICompBench \cite{huang2023t2icompbench}. We generate paraphrases and contrast prompts for each original prompt with GPT4 and generate images with Flux. Therefore, contrast prompts are more unexpected. On a small human-annotated subset, humans perform above random ($0.17$) on the inverse set, while the tested metrics do not (e.g., AlignScore with $-0.26$). This created our hypothesis that metrics might have a bias to rate unexpected matching pairs higher than contrasting pairs with a natural prompt. We test this with the categories \textit{entity placement} and \textit{entity variation} in our unsupervised dataset.}
  \label{tab:t2icompbench}
\end{table}

\section{Categories of CROC$^{hum}$}
\label{app:human_categories}
Table \ref{tab:propertiesHumanSupervised} gives an overview of the 8 selected categories for CROC$^{hum}$. Notably, this list is not exhaustive and other image generation failure cases exist, but are not covered in this work. 

\begin{table*}[t]
  \centering
  \small
  \begin{tabular}{p{0.08\textwidth} p{0.80\textwidth} }
    \toprule
    \textbf{Property} & \textbf{Description}  \\
    \midrule
    Action & The action of an entity is changed. For example, ``A ball bounces'' vs.\ ``A ball sits''.  \\
    Body Parts   & The highlighted small body part is changed. For example, ``A hand with only its ring finger colored red'' vs.\ ``A hand with only its index finger colored red''.  \\
    Counting  & The count of an entity is changed. For example, ``Two apples'' vs.\ ``Four apples''.  \\
    Negation  & One entities is negated. For example, ``A phoenix and a flag'' vs. ``A phoenix and no flag''  \\
    Shapes  & The shape of an entity is changed. For example, ``An apple in the shape of a cube'' vs.\ ``An apple in the shape of a torus''.  \\
    Size Relation  & The size relation between two objects is changed. For example ``A bigger giraffe and a smaller child'' vs.\ ``A smaller giraffe and a bigger child''.  \\
    Spatial Relation  & The spatial relation between two entities is changed. For example ``A fish left of a car'' vs.\ ``A fish right of a car''  \\
    Parts of things & The highlighted part of a thing is changed. For example, ``A bike with a blue saddle'' vs.\ ``A bike with a blue handlebar''  \\
    \bottomrule
  \end{tabular}
  \caption{Properties of CROC$^{hum}$}
  \label{tab:propertiesHumanSupervised}
\end{table*}

\section{Training Parameters for CROCScore}
We used the following training parameters: optim=adamw,  adam beta1=0.9, adam beta2=0.95, adam epsilon=1e-7, max grad norm=1.0, lr scheduler type='linear', warmup steps=100, logging steps=10, lr=$5.0e^-6$, weight decay = 0.01.
Additionally, we have saved the evaluation directions that were used in our dataset.
\label{app:TrainingParams}

\section{Templates for prompt generation}
\label{promptTemplates}

\label{app:prompt_gen_templated}
Table \ref{tab:templates} shows the templates we use for prompt generation. ``'' is copied from the first prompt. 
\ed{The prompting guides that we adapted are written by \citet{gizai} for FLUX and by \citet{sdguide} for Stable Diffusion. We chose them because of their structured breakdown and qualitative example pictures. In a second step, we further streamlined them for prompt usage in an interactive conversation with GPT-4o.}

\begin{figure}[htbp]
  \centering
  \includegraphics[width=\linewidth]{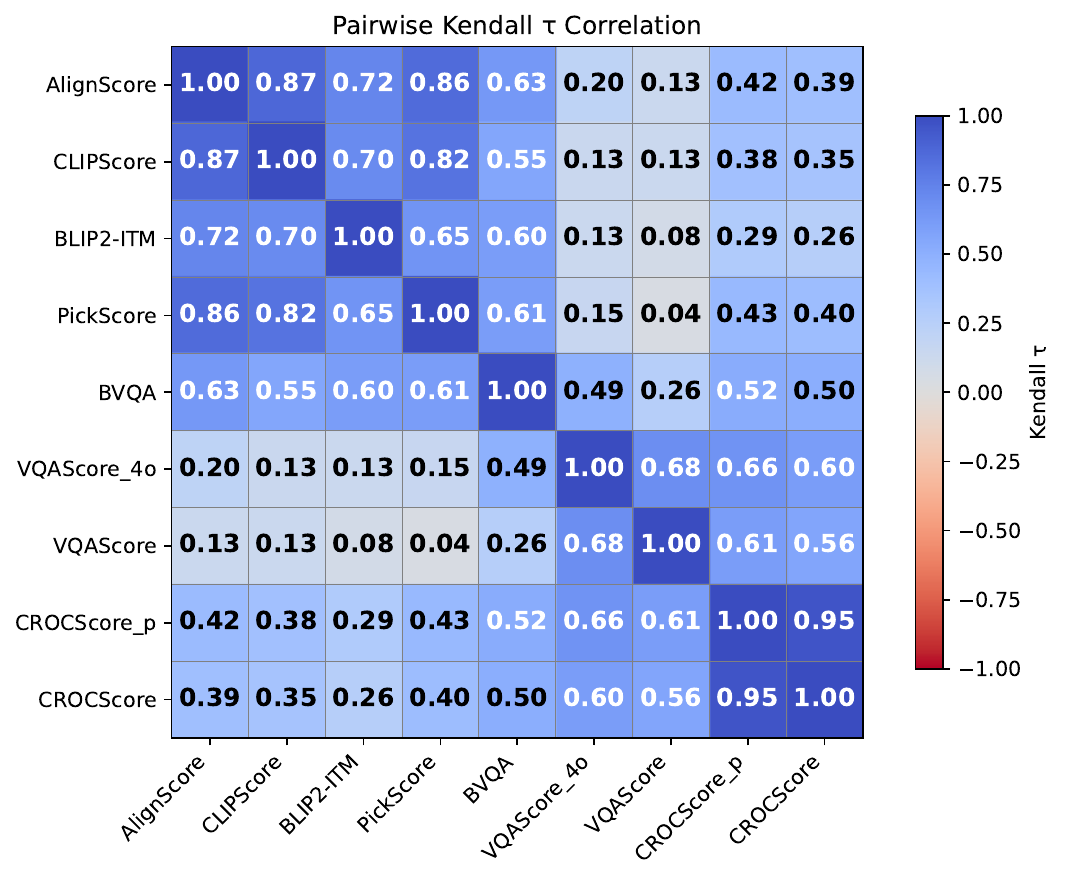}
  \caption{Kendall correlation between row-wise metric accuracies on CROC$^\text{hum}$ (Figure~\ref{fig:supervised_dataset_metric_comp}).}
  \label{fig:metriccorrs}
\end{figure}

\begin{figure}[t]
  \centering
  \includegraphics[width=\linewidth]{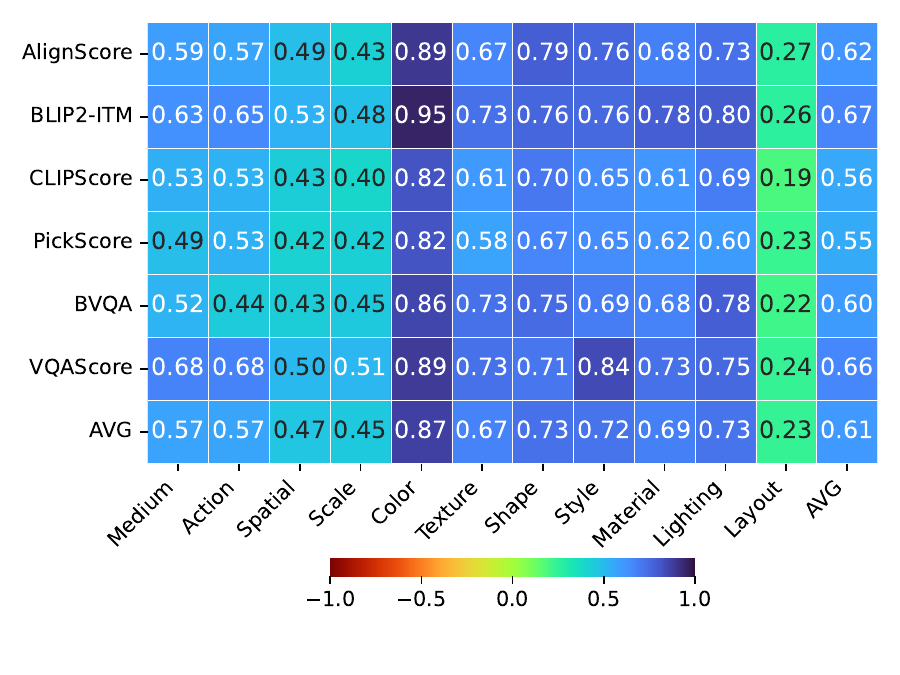}
  \medskip 
  \includegraphics[width=\linewidth]{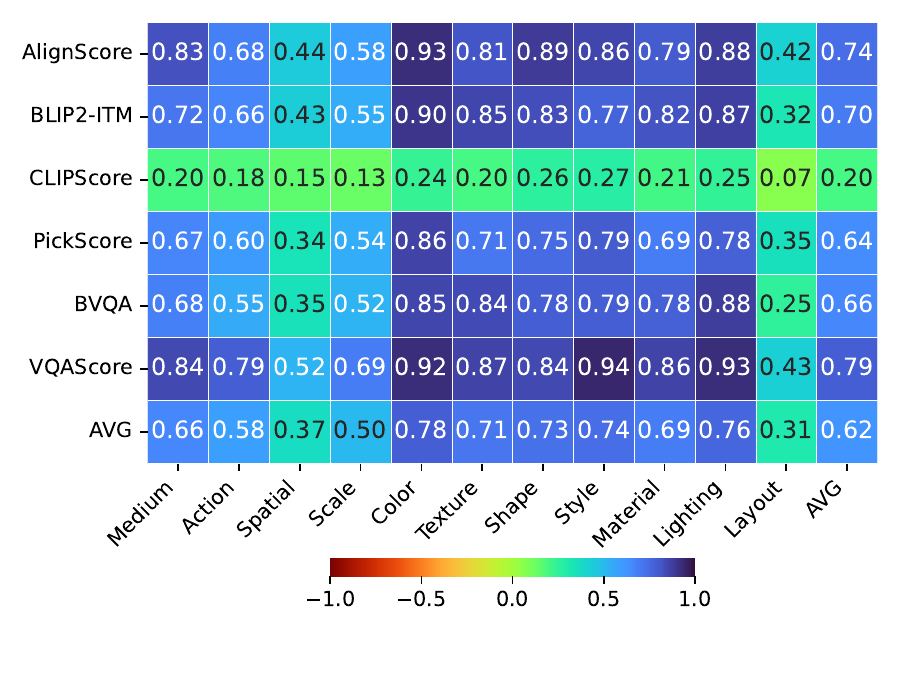}
  \caption{Scaled image-based accuracy per metric on the top-level properties of CROC$^{\text{syn}}$. \textit{Left}, samples are filtered to only include prompts smaller than or equal to 280 characters (heuristic for 77 tokens). \textit{Right}, only prompts larger than 280 characters are included. The performance of CLIPScore is degraded for long prompts (due to its token limit) and the performance of all other metrics is increased). Notably, with AlignScore outperforming BLIP2-ITM.}
  \label{fig:croc-wide}
\end{figure}

\begin{table*}[ht]
\centering
\begin{tabular}{p{0.9\textwidth}}
\hline
 \small
\begin{minipage}[t]{\linewidth}
\small
\textbf{Property Variation:} Consider the following guide on writing a good prompt with \{model\_name\}:\\
\{guide\}\\
Write a prompt for \{model\_name\} that describes a specific scene about ``\{subject\_name\}'' that involves the concept ``\{property\_name\}'' (\{property\_description\}).\\
Additionally, write a contrast prompt that strongly contrasts the original prompt in terms of the concept ``\{property\_name\}'' (\{property\_description\}), but keeps the wording and content of the prompt the same as far as possible.\\[0.5em]
For example, if the concept is a color the contrast prompt may use a different color.\\
Pay attention not to use unusual words and make sure that the contents can be displayed as images. Use simple and understandable language. Do not use phrases like ``the same'' in the contrast prompt.\\
Write the prompts very short, concise and clear. Do not write more than a single line. Do not write more than 30 words. Think step by step, then return your output in the following format:
\{\{\
\quad ``prompt'': ``Your prompt here'',\\
\quad ``contrast\_prompt'': ``Your contrast prompt here''
\}\}
\end{minipage} \\
\hline
\small
\begin{minipage}[t]{\linewidth}
\small
\textbf{Entity Variation:} '' ''\\
 Write a prompt for \{model\_name\} that describes a specific scene about ``\{subject\_name\}'' involving the entity \{entity\_name\} (Definition: \{entity\_description\}).\\
Additionally, write a contrast prompt that strongly changes parts of the entity definition \{entity\_name\} (Definition: \{entity\_description\}), but keeps the wording and content of the prompt the same as far as possible.\\
For example, if the entity is a human that has two arms, the contrast prompt may change the number of arms to three.\\[0.5em]
'' '' \{\{
\quad ``prompt'': ``Your prompt here'',\\
\quad ``varied\_definition'': ``Strongly changed definition of \{entity\_name\} (Definition: \{entity\_description\}. The definition needs to be displayable as an image and it should change the visual appearance of the entity in an unexpected way, ideally not by adding external elements, for example by changing the shape, color or changing numbers.)''\\
\quad ``contrast\_prompt'': ``Your contrast prompt here''
\}\}
\end{minipage} \\
\hline
\small
\begin{minipage}[t]{\linewidth}
\small
\textbf{Entity Placement:} '' ''\\
Write a prompt for \{model\_name\} that describes a specific scene about ``\{subject\_name\}'' with the entity \{entity\_name\} (Definition: \{entity\_description\}).\\
Additionally, write a contrast prompt that places the entity \{entity\_name\} in a picture about ``\{alt\_subject\_name\}'', but keeps the wording and content of the prompt the same as far as possible.\\
'' ''
\end{minipage} \\
\hline
\end{tabular}
\caption{Templates for Prompt Generation}
\label{tab:templates}
\end{table*}

\FloatBarrier
\end{document}